\documentclass[10pt,twocolumn,letterpaper]{article}

\usepackage[pagenumbers]{cvpr} 

\usepackage{graphicx}
\usepackage{amsmath}
\usepackage{amssymb}
\usepackage{dsfont}
\usepackage{booktabs}
\usepackage{algorithm}
\usepackage{algpseudocode}
\usepackage{multirow}
\usepackage{multicol}
\usepackage[dvipsnames,table]{xcolor}
\usepackage{siunitx}
\usepackage{pifont}%
\usepackage{tabu}
\usepackage{enumitem}
\usepackage{algorithmicx}
\usepackage{subcaption}
\usepackage{comment}
\usepackage{transparent}
\usepackage{tikz}
\usepackage[percent]{overpic}

\newcommand{\Paragraph}[1]{\vspace{0.6mm} \noindent \textbf{#1}\hspace{0mm}}
\DeclareMathOperator*{\argmax}{arg\,max}

\newcommand\norm[1]{\left\lVert#1\right\rVert}
\newcommand{\rx}[1]{#1}

\definecolor{autred}{RGB}{205, 92, 92}
\definecolor{gardengreen}{RGB}{60, 179, 113}
\definecolor{lightorange}{RGB}{255, 127, 80}

\newcommand{\boldding}[1]{%
    \rlap{\ding{#1}}%
    \kern0.2pt\ding{#1}%
}

\newcommand{\alg}{\textsc{FastMap}\xspace}
\newcommand{\coloredCell}[3]{\definecolor{mycolor}{HTML}{#2}\tikz[baseline=(char.base)]{\node[fill=mycolor,inner ysep=2pt, inner xsep=5pt, minimum width=#1, text width=#1, align=right] (char) {#3};}}

\newcommand{\cellCD}[3][24pt]{\coloredCell{#1}{#2}{#3}}

\newcommand{\tablestyle}[2]{
  \centering
  \setlength{\tabcolsep}{#1}
  \renewcommand{\arraystretch}{#2}
  \footnotesize
}

\definecolor{cvprblue}{rgb}{0.21,0.49,0.74}
\usepackage[pagebackref,breaklinks,colorlinks,citecolor=cvprblue]{hyperref}

\def\paperID{200} 
\def\confName{3DV\xspace}
\def\confYear{2026\xspace}

\title{Marginalized Bundle Adjustment: \\ Multi-View Camera Pose from Monocular Depth Estimates}

\author{Shengjie Zhu$^{1,2}$
\thanks{
Research conducted during an internship at Google. 
}  
\quad Ahmed Abdelkader$^1$  \quad Mark J. Matthews$^1$ \quad Xiaoming Liu$^2$ \quad Wen-Sheng Chu$^1$
\\[1ex]
$^1$Google \quad $^2$Michigan State University \vspace{-1ex}
}

\begin{document}
\maketitle
\begin{abstract}
Structure-from-Motion (SfM) is a fundamental 3D vision task for recovering camera parameters and scene geometry from multi-view images. 
While recent deep learning advances enable accurate Monocular Depth Estimation (MDE) from single images without depending on camera motion, integrating MDE into SfM remains a challenge.
Unlike conventional triangulated sparse point clouds, MDE produces dense depth maps with significantly higher error variance.
Inspired by modern RANSAC estimators, we propose Marginalized Bundle Adjustment (MBA) to mitigate MDE error variance leveraging its density. 
With MBA, we show that MDE depth maps are sufficiently accurate to yield SoTA or competitive results in SfM and camera relocalization tasks.
Through extensive evaluations, we demonstrate consistently robust performance across varying scales, ranging from few-frame setups to large multi-view systems with thousands of images.
Our method highlights the significant potential of MDE in multi-view 3D vision.
Code is available at \url{https://marginalized-ba.github.io/}.
\end{abstract}

\vspace{-1ex}
\section{Introduction}


Structure-from-Motion (SfM) is a fundamental method in 3D vision for recovering 3D scene geometry (as point clouds) and camera parameters (intrinsics and extrinsics) from multi-view images. 
Its versatility has fueled a wide range of applications, including 3D reconstruction \cite{furukawamulti}, robot navigation \cite{gul2019comprehensive}, camera re-localization \cite{glocker2013real}, neural rendering \cite{mildenhall2021nerf}, etc. 
Classical SfM pipelines operate by identifying sparse 2D correspondences from image pairs to jointly optimize 3D point positions and camera poses via Bundle Adjustment (BA). 
However, reliance on explicit feature matching makes these systems prone to failure in scenes with low texture or limited parallax, and thus compromises reconstruction accuracy.


Unlike classic SfM that relies on motion cues to infer geometry, deep learning advances now enable the inference of structure {\em independently} of motion via Monocular Depth Estimation (MDE) \cite{piccinelli2024unidepth,bhat2023zoedepth}. 
Despite the availability of this rich structural prior, its integration into multi-view pipelines remains an open challenge. 
Crucially, dense MDE predictions remain underutilized.
Existing works typically use them only to initialize sparse keypoints, discarding the dense data in favor of traditional BA refinement \cite{bian2022nopenerf, smith25flowmap, wang2025vggt}. 
Meanwhile, alternative learning-based methods face other limitations. 
Scene coordinate regression methods \cite{DSAC_PAMI2022, Brachmann_2023_CVPR, brachmann2024acezero} require expensive scene-specific fine-tuning.
Methods formulating BA as network inference \cite{wang2023posediffusion, wang2024vggsfm, wang2025vggt} suffer from high memory footprints that limit scalability.
Finally, other approaches that train a depth model during BA \cite{smith25flowmap} are too memory-intensive to leverage large SoTA foundation MDE models.

\begin{figure*}[!t]
    \centering
    \begin{tikzpicture}
    \node[anchor=south west,inner sep=0] (image) at (0,0) {\includegraphics[width=0.95\linewidth]{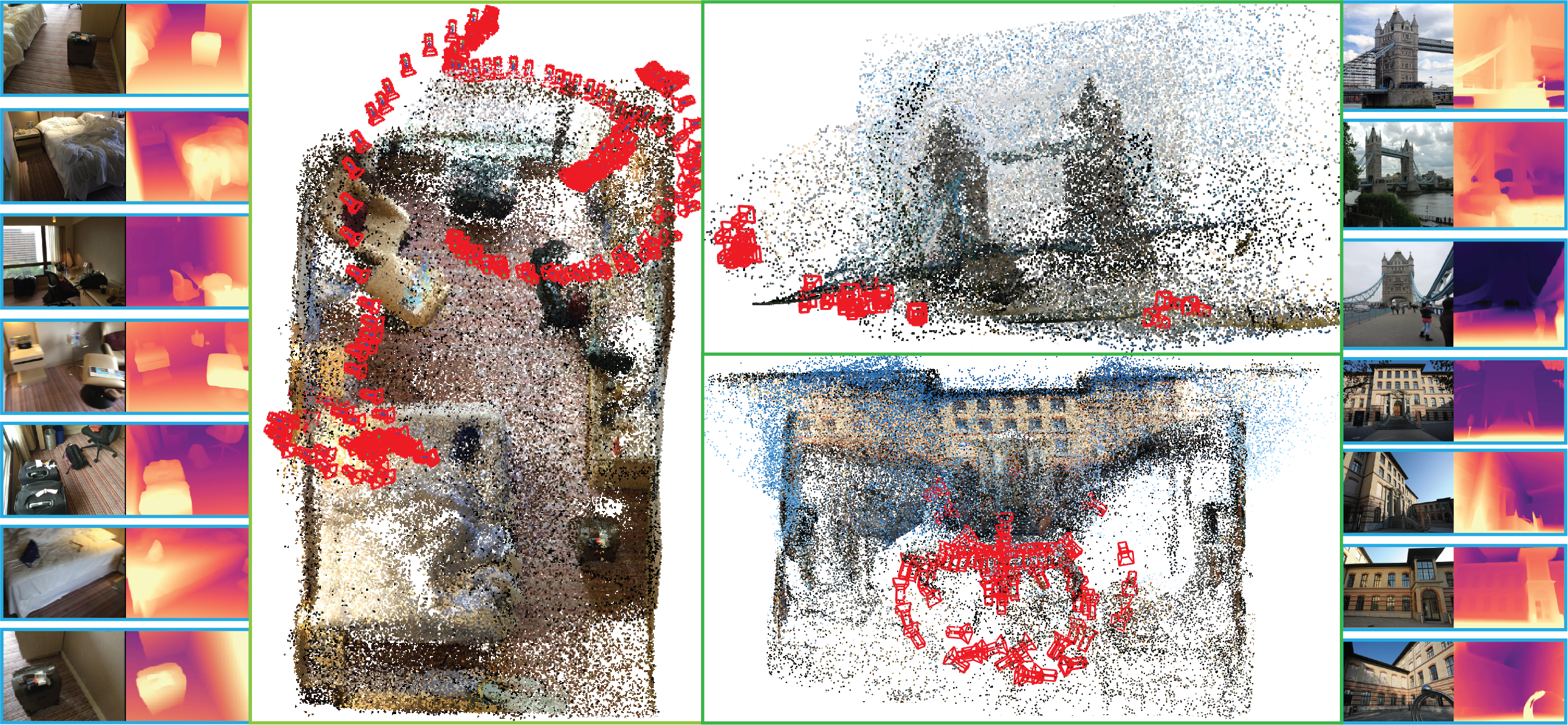}};
    \begin{scope}[x={(image.south east)},y={(image.north west)}]
        \node[fill=white, fill opacity=0.9, draw=black, text=black, font=\bfseries, inner sep=4pt] at (0.214,0.95) {\small ScanNet};
        \node[fill=white, fill opacity=0.9, draw=black, text=black, font=\bfseries, inner sep=4pt] at (0.505,0.95) {\small IMC2021};
        \node[fill=white, fill opacity=0.9, draw=black, text=black, font=\bfseries, inner sep=4pt] at (0.498,0.465) {\small ETH3D};
    \end{scope}
    \end{tikzpicture}
    \caption{
    {\bf Marginalized Bundle Adjustment (MBA).}
    Our method registers monocular depth maps into a consistent 3D coordinate system.
    Red camera icons indicate the viewpoints of registered depth maps.
    Monocular depth provides a strong structural prior, yet its predictions are inherently high-variance, as reflected by the noisy appearance of the reconstructed point cloud.
    This makes classical Bundle Adjustment designed for accurate sparse point cloud ill-suited. 
    We therefore introduce a RANSAC-inspired Bundle Adjustment objective that leverages the depth maps’ density to robustly accommodate their variance. Although depth-derived point clouds have lower visual fidelity, our experiments show that monocular depth already supports SoTA or competitive performance on diverse SfM benchmarks, as exemplified ScanNet~\cite{dai2017scannet}, IMC2021~\cite{bi2021method}, and ETH3D~\cite{bi2021method}. 
    It highlights significant potential of monocular depth models for multi-view vision tasks.
    \vspace{-3mm}}
    \label{fig:teaser}
\end{figure*}
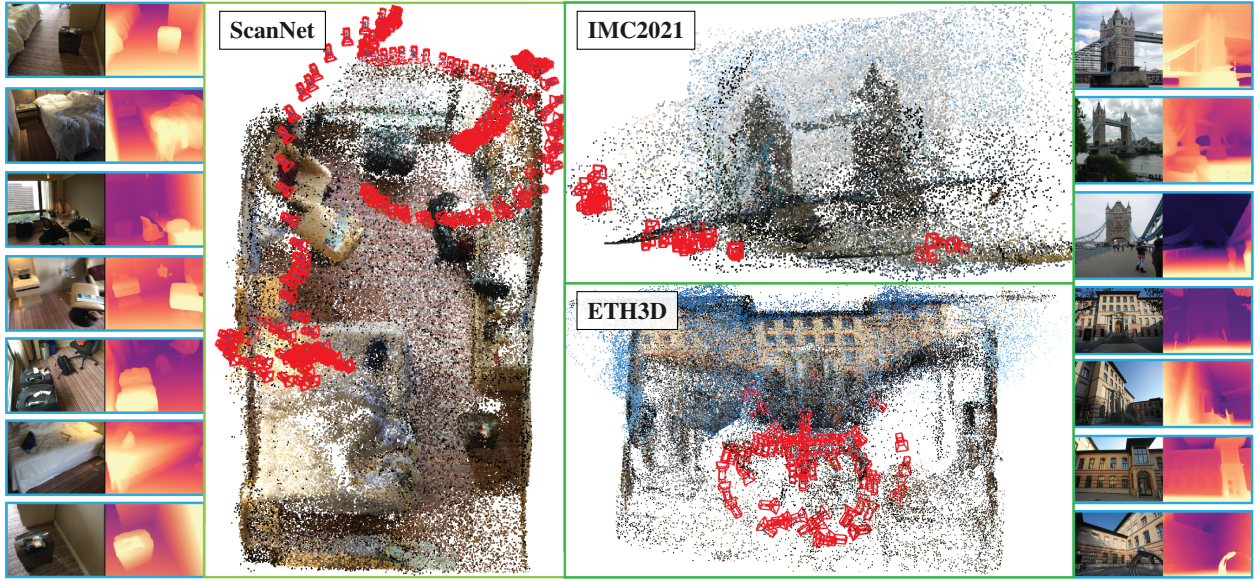

This raises a key question: \textit{how can dense MDE predictions be leveraged for multi-view pose estimation?} 
The challenge is that monocular depth maps yield dense but high-variance point clouds, failing to meet the requirements of classical SfM for sparse, accurate features (\cref{fig:teaser}). 
To bridge this gap, we propose a ``Motion-from-Structure'' approach that directly recovers camera motion from dense structural information provided by MDE.  
To faithfully showcase MDE for multi-view pose estimation, we avoid per-pixel refinement entirely, intervening only to resolve scale ambiguity through per-frame affine corrections.


To handle dense, high-variance inputs, we draw inspiration from RANSAC. 
Conventional RANSAC relies on discrete inlier counting. 
Despite being robust, it is non-differentiable and sensitive to the chosen error threshold.
To overcome this limitation, we leverage the dense projective residuals induced by MDE's dense depth predictions.  
In this setting, we observe that the inlier count for a given threshold $\tau$ corresponds (in the limit) to the Cumulative Distribution Function (CDF), $F(\tau)$, for that empirical residual distribution. 
Based on this insight, we formulate a robust BA objective that maximizes the Area Under the Curve (AUC) of this empirical CDF (\cref{fig:CDF}).
This effectively integrates information across a range of thresholds to marginalize out the error threshold in BA, hence the name Marginalized Bundle Adjustment (MBA).

\vspace{2mm}

Since analytical AUC maximization is intractable, we derive a differentiable surrogate loss.
Notably, this formulation generalizes MAGSAC~\cite{barath2019magsac}, a special case of our loss if its assumed Chi-squared distribution is replaced with the empirical residual distribution. 
Our marginalized objective is general to serve as a robust scoring function in RANSAC, where it matches the performance of MAGSAC++~\cite{barath2020magsac++} on the two-view essential matrix estimation.

We integrate the MBA objective into a flexible, coarse-to-fine framework applicable to both SfM and camera re-localization.
This pipeline is designed to accept various robust loss functions, where we opt to skip conventional components such as feature tracking and triangulation to highlight the benefits of MDE on its own.
Through the power of monocular depth priors, our results show state-of-the-art or competitive results on multiple indoor and outdoor benchmarks.
Additionally, the method proves highly scalable, capable of performing global BA over thousands of frames using a distributed cluster (Table \ref{tab:scalability}), which validates the viability of MDE-based methods for large-scale reconstruction.
We summarize our contributions as follows:
\begin{itemize}[noitemsep,topsep=0pt, leftmargin=*]
    \item First framework integrating general MDE models into SfM and re-localization tasks across varying scales.
    \item Novel and principled RANSAC-inspired objective function designed to handle dense, high-variance depth priors. This formulation is versatile and applicable to two-view RANSAC and multi-view Bundle Adjustment.
    \item SoTA or competitive performance on indoor \& outdoor, small \& large scale, SfM and re-localization benchmarks.
\end{itemize}

\begin{table*}[!t]
    \centering
    \resizebox{0.67\linewidth}{!}{%
    \begin{tabular}{l|cccccc}
       Dataset  & ScanNet~\cite{ScanNet_Dai_2017_CVPR}  & Tanks\&Temples~\cite{knapitsch2017tanks} & ETH3D~\cite{Schops_2019_CVPR} & IMC2021~\cite{bi2021method} & 7-Scenes~\cite{shotton2013scene} &
       Wayspots~\cite{Brachmann_2023_CVPR} \\
       \hline
    Images & $391$ & $1{,}106$ & $76$ & $25$ & $8{,}000$ & $1{,}157$ \\
    Pairs  & $21{,}982$ & $210{,}894$ & $1{,}571$ & $300$ & $282{,}209$ & $666{,}598$ \\
    \end{tabular}
    }
    \vspace{-2mm}
    \caption{\textbf{Maximum Size of Pose Graph} in each benchmarked dataset.
    The nodes and edges are its images and co-visible image pairs.
    \vspace{-2mm}}
    \label{tab:scalability}
\end{table*}
\section{Related Work} 
\label{sec:related}
\vspace{-1ex}

{\bf Dense Depth Map and Point Cloud Estimation.}
Recent depth foundation models~\cite{depthanything, kopf2021robust, lee2019big, birkl2023midas} enable relatively accurate metric depth maps inference in-the-wild~\cite{bhat2023zoedepth, piccinelli2024unidepth, piccinelli2025unik3d}.
Methods~\cite{wei2020deepsfm, kopf2021robust, tang2018banet} augment monocular depth with video.
Next, monocular point cloud models including MoGe~\cite{wang2024moge, wang2025moge} directly regress 3D point cloud instead of depth map, augmenting depth learning with intrinsic prior.
Further, binocular point cloud models~\cite{zhang2024monst3r, wang2025continuous} such as DUSt3R~\cite{Wang_2024_CVPR} and MASt3R~\cite{mast3r_arxiv24} integrate motion prior via regressing spatially aligned point clouds.
Yet, shown in \cref{fig:teaser}, aforementioned models are inherently dense and high-variance.
Our Marginalized BA function is designed to accommodate these characteristics. 
To our knowledge, we are also the first generalizable framework that extends depth and point cloud models to multi-view vision tasks.

\noindent {\bf RANSAC.}
RANSAC algorithms~\cite{barath2019magsac} robustly estimate low-DoF parameters under noise.
Our work extends the RANSAC philosophy from low-DoF two-view tasks to the high-DoF multi-view problems.
Namely, we handle noisy inputs from high-variance network regressions including dense depth maps. 
Several RANSAC variants~\cite{torr2000mlesac, barath2019magsac} have improved scoring function by moving from binary~\cite{fischler1981random} to continuous formulations. 
MAGSAC~\cite{barath2019magsac} smooths binary counting function by integrating residuals against a chi-squared prior distribution. 
We propose a generalized formulation that maximizes inlier counts aggregated at multiple residual thresholds. 
In essential‐matrix estimation, our strategy matches MAGSAC++'s performance~\cite{barath2019magsac}. 

\noindent {\bf Deep Learning for Multi-view Pose Estimation.}
Traditional SfM pipelines like COLMAP~\cite{Schonberger_2016_CVPR} have been improved by learned image matching models, including PixSfM~\cite{lindenberger2021pixel}, DF-SfM~\cite{He_2024_CVPR}, and Dense-SfM~\cite{lee2025densesfmstructuremotiondense}. Subsequent works incorporate 3D priors.
For example, FlowMap~\cite{smith25flowmap} formulates SfM as a self-supervised depth learning task, while MASt3R-SfM~\cite{duisterhofmast3rsfm} introduces a dedicated point cloud aggregation strategy for models like DUSt3R~\cite{Wang_2024_CVPR} and MASt3R~\cite{mast3r_arxiv24}. More recently, the entire SfM pipeline has been reformulated as an end-to-end learning problem~\cite{wang2024vggsfm, wang2023posediffusion, wang2025vggt}. However, high memory footprint typically limits these approaches to small-scale scenarios. In the related task of camera re-localization, methods often train a network to map images to world coordinates but usually require scene-specific fine-tuning~\cite{brachmann2024acezero, arnold2022map}.  To our knowledge, MBA is the first framework to successfully apply general-purpose MDE models to both small and large-scale SfM problems. Notably, when using DUSt3R as its MDE model, MBA achieves higher accuracy than the specialized MASt3R-SfM pipeline~(\cref{tab:imc}).
\section{Method} \label{sec:pe_method}

We propose an MDE-based approach for multi-view pose estimation.
As shown in \cref{fig:teaser}, our pipeline keeps the depth maps fixed, and applies only affine corrections to account for the depth scale ambiguity.
Our core contribution is a marginalized BA (MBA) objective that effectively leverages dense depth map predictions with high variance.
In this section, we present our approach in the context of SfM, and defer the minor adaptations required for camera re-localization and two-view RANSAC towards the end.

\subsection{System Overview}
\label{sec:preliminaries}
\begin{figure*}[!t]
    \centering
    \begin{tikzpicture}
    \node[anchor=south west,inner sep=0] (image) at (0,0) {
        \includegraphics[height=3.45cm]{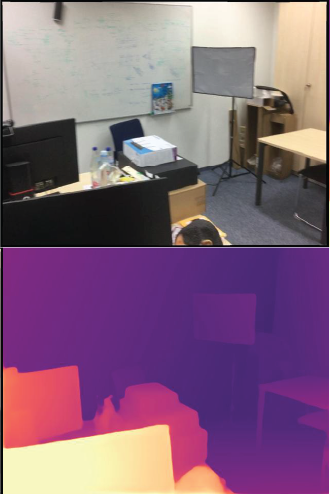}\;
        \includegraphics[height=3.45cm]{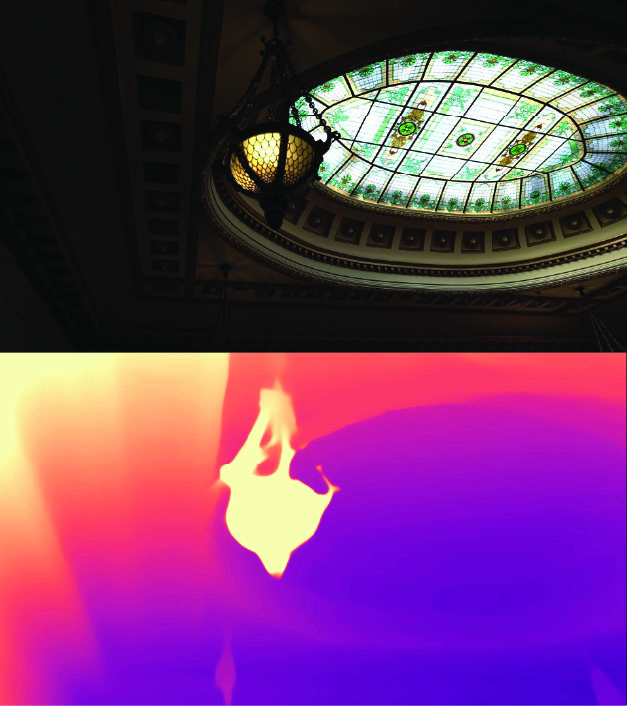}\;
        \includegraphics[height=3.45cm]{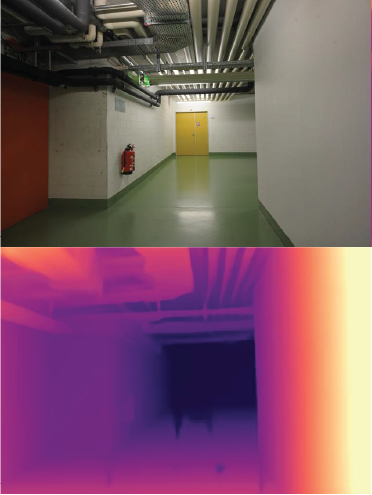}\;
        \includegraphics[height=3.45cm]{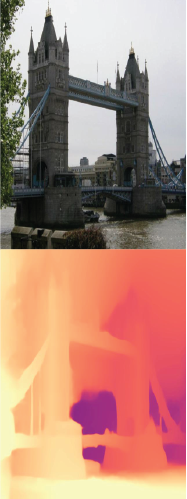}\;
        \includegraphics[height=3.45cm]{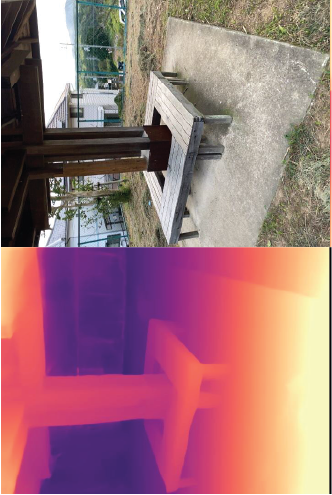}\;
        \includegraphics[height=3.45cm]{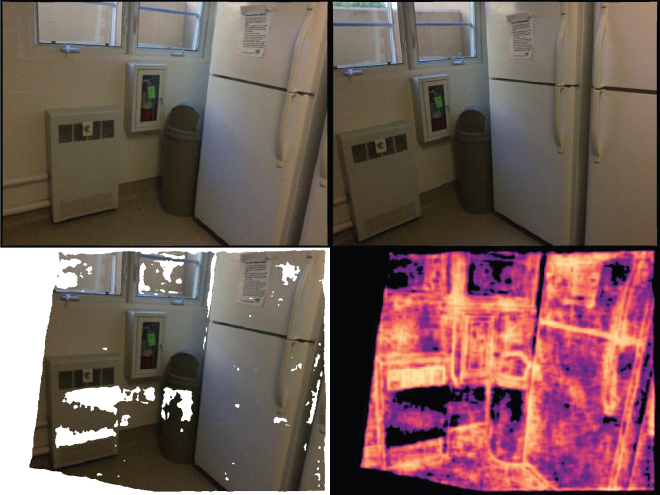}
    };
    \begin{scope}[x={(image.south east)}, y={(image.north west)}]
        \node[fill=white, fill opacity=0.8, draw=black, text=black, font=\bfseries, inner sep=2pt] at (0.042-0.005,0.92) {\tiny (a) ScanNet};
        \node[fill=white, fill opacity=0.8, draw=black, text=black, font=\bfseries, inner sep=2pt] at (0.221-0.045,0.92) {\tiny (b) T\&T};
        \node[fill=white, fill opacity=0.8, draw=black, text=black, font=\bfseries, inner sep=2pt] at (0.221+0.147,0.92) {\tiny (c) ETH3D};
        \node[fill=white, fill opacity=0.8, draw=black, text=black, font=\bfseries, inner sep=2pt] at (0.413+0.11,0.92) {\tiny (d) IMC};
        \node[fill=white, fill opacity=0.8, draw=black, text=black, font=\bfseries, inner sep=2pt] at (0.534+0.086,0.92) {\tiny (e) Wayspots};
        \node[fill=white, fill opacity=0.8, draw=black, text=black, font=\bfseries, inner sep=2pt] at (0.730+0.05,0.92) {\tiny (f1) Source Image};
        \node[fill=white, fill opacity=0.8, draw=black, text=black, font=\bfseries, inner sep=2pt] at (0.895+0.02,0.92) {\tiny (f2) Target Image};
        \node[fill=white, fill opacity=0.8, draw=black, text=black, font=\bfseries, inner sep=2pt] at (0.737+0.048,0.42) {\tiny (f3) Correspondence};
        \node[fill=white, fill opacity=0.8, draw=black, text=black, font=\bfseries, inner sep=2pt] at (0.917+0.015,0.42) {\tiny (f4) Confidence of Corres.};
    \end{scope}
    \end{tikzpicture}
    \caption{\textbf{Dense Depth and Correspondence} are inputs to our multi-view pose estimation system. 
     We benchmark pose performance with depth map under diverse imaging conditions. 
     Figures (a) to (e) include ScanNet~\cite{ScanNet_Dai_2017_CVPR} indoor images, T\&T~\cite{knapitsch2017tanks} and ETH3D~\cite{Schops_2017_CVPR} high resolution images, IMC2021~\cite{bi2021method} internet-collected images, and Wayspots~\cite{Brachmann_2023_CVPR} flipped image.
     Fig.~(f) visualizes dense correspondence.
     \vspace{-2mm}}
    \label{fig:inputs}
\end{figure*}
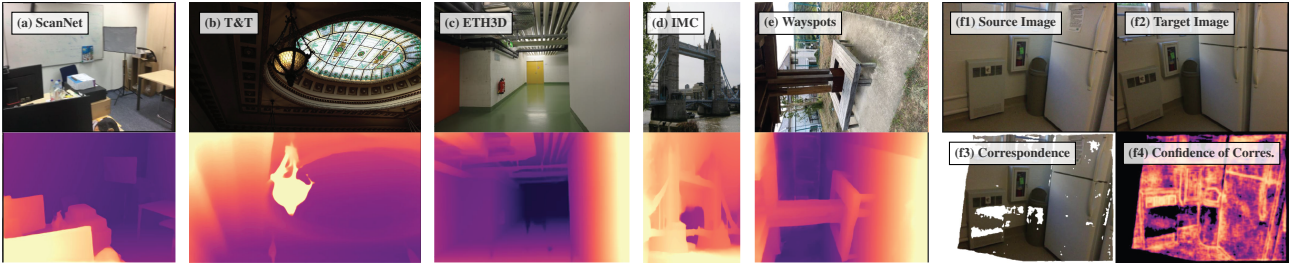

\Paragraph{Problem Definition.} 
Given as input an unordered collection of $N$ RGB frames $\{\mathbf{I}_i\}_{i \in [N]}$, we optimize for camera intrinsics $\mathcal{K} \!=\! \{ \mathbf{K}_i\}$ and extrinsics $\mathcal{P} \!=\! \{\mathbf{P}_i\}$.  To do so, we precompute $N$ depth maps $\mathcal{D} = \{\mathbf{D}_i  \!=\! \mathcal{N}_{\mathbf{D}}(\mathbf{I}_i)\}$ and pairwise correspondence maps $\mathcal{C} = \{\mathbf{C}_{i, j} \!=\! \mathcal{N}_{\mathbf{C}}(\mathbf{I}_i, \mathbf{I}_j), i \neq j\}$, where $\mathcal{N}_{\mathbf{D}}$ and $\mathcal{N}_{\mathbf{C}}$ denote pre-trained depth and correspondence models, respectively (see examples in \cref{fig:inputs}).
We jointly optimize per-frame affine corrections $\mathcal{A} = \{\alpha_i, \beta_i \mid i \leq N \}$ for each depth map, 
producing scale-ambiguity corrected depth maps:
\begin{equation}
\mathbf{D}_i' = \alpha_i \cdot \mathbf{D}_i + \beta_i.
\label{eqn:depth_adjustment}
\end{equation}

\Paragraph{Optimization Objective.} 
Denote $\mathcal{X} = \{\mathcal{P}, \mathcal{K}, \mathcal{A}\}$ as the set of all variables to optimize, and $\mathbf{X}_i = \bigl(\mathbf{P}_i, \mathbf{K}_i, \mathbf{A}_i\bigr)$, we define the objective as maximizing a scoring function $\mathcal{S}$:
\begin{equation}
    \mathcal{X}^* = \argmax_{\mathcal{X} = (\mathcal{P}, \mathcal{K}, \mathcal{A})} \mathcal{S}(\mathcal{X} \mid \mathcal{D}, \mathcal{C}).
    \label{eqn:overview_loss}
\end{equation}
We define $\mathcal{S}$ as the summation of a suitable quality function $\mathcal{Q}$ over frame pairs $(I_i, I_j)$ over the pose graph $\mathcal{G}$:
\begin{equation}
    \mathcal{S}(\mathcal{X} \mid \mathcal{D}, \mathcal{C}) = \sum_{(i, j) \in \mathcal{G}} \mathcal{Q}(\mathbf{X}_i, \mathbf{X}_j\mid \mathbf{D}_i, \mathbf{D}_j, \mathbf{C}_{i, j}).
    \label{eqn:pair_loss_1}
\end{equation}
The pose graph $\mathcal{G}$ (defined in \cref{sec:sfm_pipeline}) connects co-visible frame pairs. 
$Q(\cdot)$ is realized with various forms in \cref{sec:ba}.

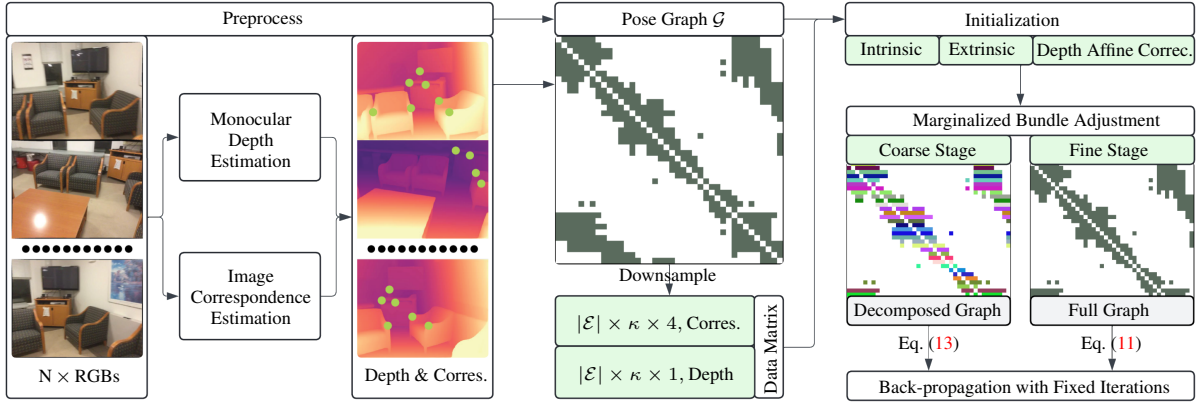
\begin{figure*}[t]
  \centering
  \begin{overpic}[width=0.92\linewidth,grid=false,percent]{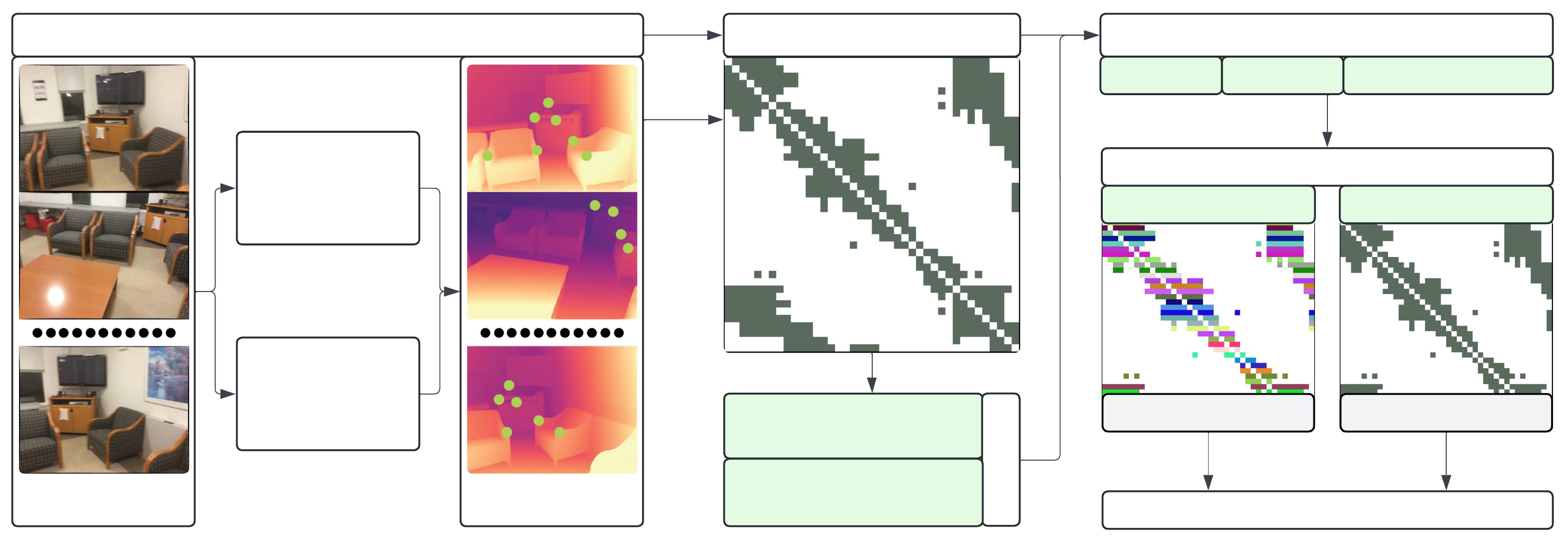}
    \put(6.8, 2.85){\makebox(0,0){\scriptsize N\,$\times$\,RGBs}}
    
    \put(35.5, 2.85){%
      \makebox(0,0){%
        \parbox{2.0cm}{\centering\scriptsize Depth \& Corres.
            }%
        }
    }

    \put(22.0, 32.4){%
      \makebox(0,0){%
        \parbox{1.5cm}{\centering\scriptsize Preprocess
            }%
        }
    }
    \put(21.0, 22.5){%
      \makebox(0,0){%
        \parbox{1.7cm}{\centering\scriptsize Monocular Depth Estimation
            }%
        }
    }

    \put(21.0, 9.5){%
      \makebox(0,0){%
        \parbox{1.7cm}{\centering\scriptsize Image Correspondence Estimation
            }%
        }
    }
    
    \put(55.5, 11.2){\makebox(0,0){\scriptsize Downsample}}
    \put(56.0, 32.4){\makebox(0,0){\scriptsize Pose Graph $\mathcal{G}$}}
    \put(55.0, 7.2){%
      \makebox(0,0){%
        \parbox{3.0cm}{\centering\scriptsize 
        $| \mathcal{E}|\times \kappa \times 4, \text{Corres.}$
            }
        }
    }
    \put(54.5, 3.0){\makebox(0,0){\scriptsize         $| \mathcal{E}|\times \kappa \times 1, \text{Depth}$}}
    \put(64, 5.2){\rotatebox{90}{\makebox(0,0){\scriptsize Data Matrix}}}

    \put(84,32.4){\makebox(0,0){\scriptsize Initialization}}
    \put(74,30){\makebox(0,0){\scriptsize Intrinsic}}
    \put(81.5,30){\makebox(0,0){\scriptsize Extrinsic}}
    \put(92.5,29.8){\makebox(0,0){\scriptsize Depth Affine Correc.}}
    \put(86, 24){\makebox(0,0){\scriptsize Marginalized Bundle Adjustment}}
    \put(77, 21.5){\makebox(0,0){\scriptsize Coarse Stage}}
    \put(92, 21.5){\makebox(0,0){\scriptsize Fine Stage}}
    \put(77, 8.3){\makebox(0,0){\scriptsize Decomposed Graph}}
    \put(92, 8.3){\makebox(0,0){\scriptsize Full Graph}}
    \put(92.2, 5.5){\makebox(0,0){\scriptsize \cref{eqn:fine_backward}}}
    \put(77.1, 5.5){\makebox(0,0){\scriptsize \cref{eqn:coarse_ba}}}
    \put(85, 1.85){\makebox(0,0){\scriptsize Back-propagation with Fixed Iterations}}

    \put(135,78){\makebox(0,0){\scriptsize Fine Stage}}
    \put(135, 35){\makebox(0,0){\scriptsize Full Graph}}
  \end{overpic}
  \vspace{-1ex}
  \caption{%
    \textbf{System Overview.}
    With $N$ RGBs, the system consumes dense depth maps and pairwise correspondence inferred by pretrained models.
    The system outputs intrinsics, extrinsics, and frame-wise depth affine corrections scalars.
    A sparse $N\times N$ pose graph is built from co-visible frames using correspondences.
    Dense inputs are subsampled into a data matrix of $| \mathcal{E}| \times \kappa \times 5$ (graph edges count) for scalable multi-GPU optimization.
    After initialization, the Bundle Adjustment proceeds from coarse to fine. 
    In coarse stage, the BA objective is evaluated and summed over “star-shaped” subgraph $\mathcal{G}_i$ of each frame $i$.
    One subgraph includes itself plus its co‑visible neighbors, marked as one colored row in coarse pose graph.
    Fine stage computes with full graph.
    The BA applies gradient descent for fixed iterations.
    }
  \label{fig:pipeline}
\end{figure*}

\Paragraph{Algorithm Pipeline.} 
In \cref{fig:pipeline}, we subsample dense depth and correspondence into a data matrix.
Outlined in \cref{sec:sfm_pipeline}, with initialized intrinsics, extrinsics, and depth affine corrections,
we sequentially execute coarse and fine stage SfM.
Two stages are mostly same except for employing different pose graphs.
Each stage applies gradient descent with fixed iterations.
In the following \cref{sec:ba}, we first describe our core algorithm: the marginalized BA objective for dense, high-variance network regressions.

\subsection{Marginalized Bundle Adjustment (MBA)}
\label{sec:ba}

We begin with a naive yet robust binary quality function to realize the \cref{eqn:pair_loss_1} quality function.
Despite its robustness, the binarized function is non-differentiable and sensitive to the single threshold selected.
Hence, we propose a smooth form integrating over multiple thresholds.

\Paragraph{Subsample Dense Depth and Correspondence.}
For each co-visible frame pair of the pose graph $\mathcal{G}$, we sample a fixed number $\kappa$ of paired pixels.  Specifically, between frame $i$ and $j$, we sample $\kappa$ depth pixels on frame $i$ and $\kappa$ $i$-to-$j$ correspondence pixels.  
We apply random sampling with replacement over the correspondences with a confidence score at least $\chi$ yielding a data matrix of size $|\mathcal{E}| \times \kappa \times 5$, where $\mathcal{E}$ denotes the edges of $\mathcal{G}$.

\Paragraph{Projective Residuals.}
We define the residual $r_{i, j, k}$ as the 2D discrepancy in the $k^\mathrm{th}$ sampled correspondence $c_{i,j,k} \in \mathbf{C}_{i, j}$.  Denoting $c_{i,j,k}$ as $(p_{i,j,k}, q_{i,j,k}) \in I_i \times I_j$, we write
\begin{equation}
    r_{i, j, k} = \norm{\pi_{i \to j}\bigl(\mathbf{D}_i'[p_{i,j,k}]\bigr) - q_{i,j,k}}_2,
    \label{eqn:residual_2d}
\end{equation}
where the operator \( \pi_{i \to j} \) projects the pixel $p_{i,j,k}$ in frame $I_i$ to the frame \( I_j \) with its affine-aligned depth value in $\mathbf{D}_i'$ from \cref{eqn:depth_adjustment}. 
Pixel $q_{i,j,k}$ is the corresponding pixel of $p_{i,j,k}$.
The projection is defined by the camera intrinsics \( \mathbf{K}_i \), \( \mathbf{K}_j \) and extrinsics \( \mathbf{P}_i \), \( \mathbf{P}_j \)~\cite{hartley2003multiple}.  Other robust norms may also be used in \cref{eqn:residual_2d}, \textit{e.g.}, the Cauchy function used in \cite{schoenberger2016sfm}.

\Paragraph{Binary Quality with a Threshold.}
We start by realizing  \cref{eqn:pair_loss_1} with a robust binary quality function:
\begin{align}
    \mathcal{S}^b(\mathcal{X} \mid \mathcal{D}, \mathcal{C}, \tau) &=  \sum_{(i, j) \in \mathcal{G}} \left(\sum_{k \in \kappa }\mathds{1}[r_{i, j, k} < \tau] \right),
    \label{eqn:binary_scoring_funciton}
\end{align}
where the variable $\tau$ is the residual threshold and function $\mathds{1}(\cdot)$ is the indicator function.
Intuitively, a depth pixel is considered an inlier if its projective residual is below the threshold $\tau$.
The binarized scoring function in \cref{eqn:binary_scoring_funciton} is widely used in RANSAC algorithms~\cite{fischler1981random} for its robustness on overly-sampled noisy inputs.
RANSAC algorithm mostly addresses low Degree-of-Freedom (DoF) problem including essential matrix estimation~\cite{nister2004efficient}.
In contrast, multi-view pose estimation problem solves multiple poses, possessing a significantly larger solution space.
This necessitates a \textit{continuous} scoring function to enable iterative optimization whereas $\mathcal{S}^b$ from \cref{eqn:binary_scoring_funciton} is discrete.

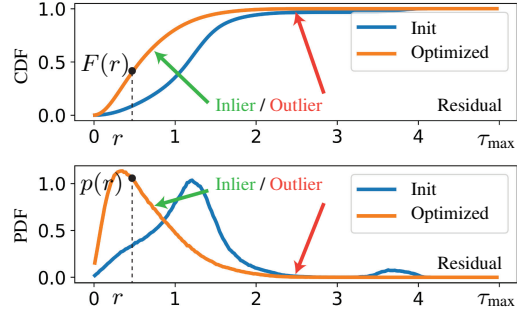
\begin{figure}[!t]
     \centering
     \begin{overpic}[width=0.8\linewidth,grid=false,percent]{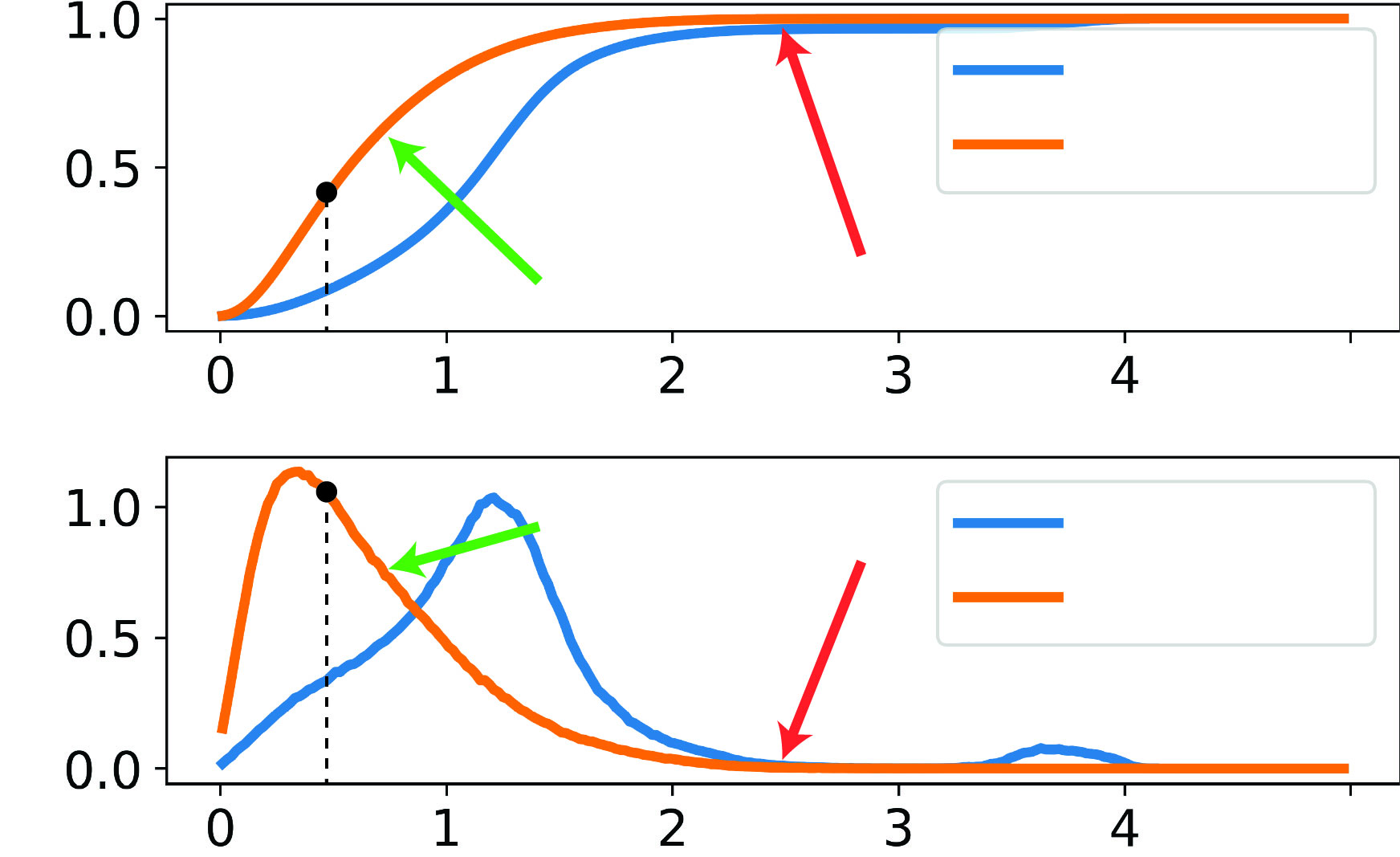}
    \end{overpic}
    \put(-40, 104){\scriptsize Init}
    \put(-40, 95){\scriptsize Optimized}
    \put(-40, 43){\scriptsize Init}
    \put(-40, 34){\scriptsize Optimized}
    \put(-30, 15)
    {\scriptsize Residual}
    \put(-30, 75){\scriptsize Residual}
    \put(-15, 2){{\small $\tau_{\text{max}}$}}
    \put(-15, 62){{\small $\tau_{\text{max}}$}}
    \put(-153, 62){{\small $r$}}
    \put(-153, 2){{\small $r$}}
    \put(-165, 90){{\small $F(r)$}}
    \put(-165, 45){{\small $p(r)$}}
    \put(-190, 25){\rotatebox{90}{\scriptsize PDF}}
    \put(-190, 85){\rotatebox{90}{\scriptsize CDF}}
    \put(-114, 45){{\scriptsize \textcolor[HTML]{39B54A}{Inlier} / \textcolor[HTML]{ef4136}{Outlier} }}
    \put(-114, 75){{\scriptsize \textcolor[HTML]{39B54A}{Inlier} / \textcolor[HTML]{ef4136}{Outlier} }}
     \caption{
     \textbf{CDF and PDF} of empirical residual distribution $\mathcal{R}$. 
     In \cref{eqn:binary_scoring_funciton_thresholds}, our BA maximizes the area-under-the-curve of $\mathcal{R}$'s CDF curve up a maximum threshold.
     The BA automatically formulates a smoothed categorization of inliers versus outliers. 
     The forward and backward computation ( Eqn.~\ref{eqn:fine_forward} \& \ref{eqn:fine_backward} ) at the residual $r$ is succinctly defined as indexing the curve at $F(r)$ and $p(r)$.
     \vspace{-2mm}}
     \label{fig:CDF}
\end{figure}

\Paragraph{CDF as Smoothed Quality.}
The dense depth maps provide enough samples of projective residuals to utilize their distributional properties. 
Let $R$ denote the set of all residuals.
We model the residual $r$ as a random variable following an empirical distribution $\mathcal{R}$ estimated with kernel density estimation (KDE)~\cite{KDE_Sliverman_1986}.
This gives residual $r$ distribution $\mathcal{R}$:
\begin{equation}
    r \sim \mathcal{R} = \mathrm{KDE}(R), \; R= \{r_{i,j,k} \mid (i, j) \in \mathcal{G}, [k] \in \kappa\}.
\end{equation}
Please see \cref{fig:CDF} for an example distribution.
Denote \( p(r) \) and \( F(\tau) = \mathrm{Pr}[r \!<\! \tau] \) as the probability distribution function (PDF) and cumulative distribution function (CDF) of \( \mathcal{R} \), respectively.
The binary scoring function at threshold $\tau$ in \cref{eqn:binary_scoring_funciton} is approximated as the CDF function at value $\tau$. 
\begin{align}
    & \mathcal{S}^b(\mathcal{X} \mid \mathcal{D}, \mathcal{C}, \tau) =\sum_{i,j,k} \mathds{1}[r_{i, j, k} < \tau]  \approx |R| \cdot F(\tau). 
    \label{eqn:CDF_function}
\end{align}

\Paragraph{Marginalizing Thresholds.}
We generalize \cref{eqn:binary_scoring_funciton} by extending from single to multiple thresholds. 
We integrate \cref{eqn:binary_scoring_funciton} up to a maximum threshold $\tau_\text{max}$, beyond which a residual is safely considered an outlier. 
\begingroup
\begin{align}
    \mathcal{S}^m(\mathcal{X} \mid \mathcal{D}, \mathcal{C}) & = \int_{0}^{\tau_\text{max}} \mathcal{S}^b(\mathcal{X} \mid \mathcal{D}, \mathcal{C}, \tau) \, d \tau \notag \\ & \approx |R| \,  \int_{0}^{\tau_\text{max}} F(r) \, d \tau.
    \label{eqn:binary_scoring_funciton_thresholds}
\end{align}
\endgroup
Intuitively, depth pixels from dense depth maps vary at noise levels.
In \cref{eqn:binary_scoring_funciton}, a large threshold \( \tau \) aids to register camera at approximately correct locations while a small \( \tau \) improves accuracy but risks local minima.
On the other hand, \cref{eqn:binary_scoring_funciton_thresholds} leverages benefits from setting both small and large thresholds.
Mathematically, it computes the area under a truncated CDF curve, as outlined in \cref{fig:CDF}.

\Paragraph{Marginalized Bundle Adjustment (MBA) Objective.}
Practically, we use a histogram-based KDE implementation. 
This renders \cref{eqn:binary_scoring_funciton_thresholds} as a summation over a finite set of $T$ thresholds with $\tau_i = \frac{i}{T} \cdot \tau_{\max} $.
\begin{align}
    \mathcal{S}^m(\mathcal{X} \mid \mathcal{D}, \mathcal{C})
    &= \sum_{ 0 \leq i\ \leq T} \mathcal{S}^b(\mathcal{X} \mid \mathcal{D}, \mathcal{C}, \tau_i) \notag \\
    & \approx |R| \cdot \frac{\tau_{\max}}{T} \cdot \sum_{ i = 0}^{T}  F(\tau_i).
    \label{eqn:binary_scoring_funciton_thresholds_summation}
\end{align}

\Paragraph{MBA Forward.} 
In \cref{fig:CDF}, to maximize \cref{eqn:binary_scoring_funciton_thresholds_summation}, we propose a \textit{surrogate loss} with pixel-wise forward and backward functions (plus a negative sign).
Intuitively, \cref{eqn:binary_scoring_funciton_thresholds_summation} provides a smooth differentiable loss wrapped over each residual:
\begin{align}
    \mathcal{L}_{\text{MBA}}
    = \frac{1}{|R|} \sum_{i, j, k} -F(r_{i, j, k}) \cdot \mathds{1}[r_{i, j, k} < \tau_{\text{max}}] .
    \label{eqn:fine_forward}
\end{align}

\Paragraph{MBA Backward.} 
Correspondingly, the backward is:
\begin{align}
    \frac{\partial \mathcal{L}_{\text{MBA}}}{\partial r_{i, j, k} }
    = -\frac{1}{|R|} p(r_{i, j, k} ) \cdot \mathds{1}[r_{i, j, k} < \tau_{\text{max}}].
    \label{eqn:fine_backward}
\end{align}

\Paragraph{Discussion.} 
Despite employing \cref{eqn:fine_forward} as loss, our optimization objective is \cref{eqn:binary_scoring_funciton_thresholds_summation}.
We prove its effectiveness via applying \cref{eqn:binary_scoring_funciton_thresholds_summation} as a scoring function to a two-view pose estimation with RANSAC.
In \cref{tab:ransac_combined}, our scoring function performs comparable to SoTA method \cite{barath2019magsac} with dense image correspondence inputs.
Notably, MAGSAC~\cite{barath2019magsac} scoring function becomes a special case of \cref{eqn:fine_forward} by replacing its assumed chi-squared distribution with the empirical distribution $\mathcal{R}$. 
See more details in Supp.~Sec.~\ref{sec:surrogate_loss_to_magsac}.
\paragraph{Robustness.}
As a RANSAC inspired objective function, the proposed loss function \cref{eqn:binary_scoring_funciton_thresholds_summation} inherits its robustness.
From the backward function \cref{eqn:fine_backward}, the gradient of extreme residual values, \textit{i.e.}, those of low probability, is suppressed, as shown in \cref{fig:CDF}.
After convergence, the noise level of each depth pixel is implicitly captured by its residual probability, enabling BA to automatically distinguish inliers from outliers without a dedicated neural network. 

\Paragraph{Scalability.} 
Shown in \cref{fig:pipeline}, our approach maintains an $| \mathcal{E}| \times \kappa \times 5$ matrix. It allows scaling up by parallelizing computation across multiple GPUs. In \cref{tab:scalability}, we enable global BA over large-scale pose graph involving $8,000$ frames and $564,418$ co-visible frame pairs. 
This significantly surpasses recent methods including FlowMap~\cite{smith25flowmap}, VGG-SfM~\cite{wang2024vggsfm}, and PoseDiffusion~\cite{wang2023posediffusion}, which run out of memory when processing more than $200$ frames~\cite{duisterhofmast3rsfm}.




\subsection{SfM Pipeline}
\label{sec:sfm_pipeline}
The section outlines the sequentially executed, initialization, coarse-stage SfM, and fine-stage SfM processes.
We also describe extension to camera re-localization.

\Paragraph{Pose Graph Construction.} 
As shown in \cref{fig:pipeline}, we construct an undirected pose graph $\mathcal{G}$ from a set of co-visible frame pairs $(I_i, I_j)$. 
Co-visibility between two frames is measured as the percentage of pixels that are visible in both views, computed from correspondence maps $\mathbf{C}_{i, j} \in \mathcal{C}$. 
We include an edge $g_{i,j} \in \mathcal{G}$ if the co-visibility score exceeds a threshold $\nu$, resulting in a sparse pose graph.

\Paragraph{Intrinsic Initialization.} 
Each frame, we run DUSt3R~\cite{Wang_2024_CVPR} with two identical frames to extract a dense point cloud, then initialize per-frame intrinsics via RANSAC-based calibration~\cite{zhu2023tame}. 
If a shared intrinsic is assumed, we initialize it using the median focal length.
See details at Supp.~Sec.~\ref{sec:supp_details}.

\Paragraph{Camera Pose and Depth Adjustment Initialization.}
We construct a spanning tree from the pose graph $\mathcal{G}$ using a greedy strategy. 
The root node is chosen as the frame with the highest degree. 
At each step, we add a new node that maximizes the total degree of the tree. For each new frame $i$, we identify a registered, co-visible frame $j$. 
Using projection from frame $j$ to $i$, we initialize the 6-DoF camera pose $\mathbf{P}_i$, with translation magnitude estimated following a Five-Point algorithm-based method. 
From the reverse projection $i$ to $j$, we estimate depth scale adjustment $s_i$. 
The depth bias $\beta_i$ is initialized to zero.

\begin{figure}[!t]
    \centering
    \subfloat[7-Scenes~\cite{shotton2013scene}, sequence Stairs]{%
        \includegraphics[width=0.49\linewidth]{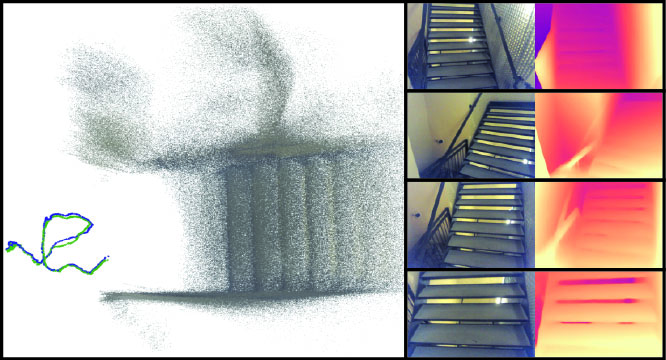}%
        \label{fig:camreloc-a}%
    }\hfill
    \subfloat[Wayspots~\cite{Brachmann_2023_CVPR}, sequence Lawn]{%
        \includegraphics[width=0.49\linewidth]{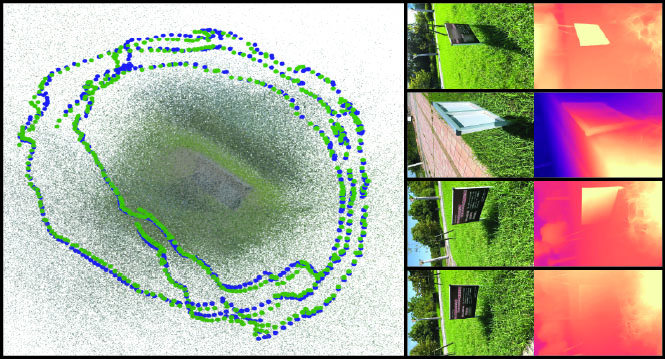}%
        \label{fig:camreloc-b}%
    }
    \caption{%
        {\bf Camera Re-localization} on 7-Scenes and Wayspots. 
       \textcolor{ForestGreen}{Green} and \textcolor{blue}{blue} mark \textcolor{ForestGreen}{predicted} and \textcolor{blue}{groundtruth} odometry. We present challenging sequences of repetitive, textureless images. The images exhibit (1) scale changes, (2) flipping, and (3) a lack of distinguishable depth references. Consequently, the model predicts sub-optimal depth maps. Surprisingly, despite these significant challenges, depth maps still support accurate camera poses.
    \vspace{-2mm}}
    \label{fig:camreloc}
\end{figure}

\Paragraph{Coarse-Stage SfM.}
Suppose the frame $i$ is poorly registered, its corresponding residual \( r_{i, j, k} \) exhibits significantly large values.
Due to the robustness property of \cref{eqn:fine_backward}, the residuals with larger values are automatically suppressed with smaller gradients, see outliers at \cref{fig:CDF}.
These characteristics cause poorly registered frames to become ``stuck" in a local minimum.
We propose a graph decomposition strategy to mitigate the occurrence of early local minima.
For the graph $\mathcal{G}$, we decompose it into a $N$ subgraphs $\mathcal{G}_i$:
\begin{align}
    \mathcal{G} = \sum_{i \in N} \mathcal{G}_i, \; \; \mathcal{G}_i = \{\mathcal{V}_i, \; \mathcal{E}_i\},
\end{align}
where $\mathcal{V}_i \!=\! \{ \mathbf{I}_i \} \cup \mathcal{N}(\mathbf{I}_i)$  and $ \mathcal{E}_i \!=\! \{ (\mathbf{I}_i, \mathbf{I}_j) \mid \mathbf{I}_j \in \mathcal{N}(\mathbf{I}_i) \}$.
Each subgraph $\mathcal{G}_i$ consists of the $i$-th frame $\mathbf{I}_i$ and its neighboring frames $\mathcal{N}(\mathbf{I}_i)$. 
In \cref{fig:pipeline}, we visualize each subgraph as a row of color in the sparse pose graph.

Beyond subgraph decomposition, we apply logarithmic operation to the L2 norm residual in \cref{eqn:residual_2d} to enhance its robustness to large residuals from poorly registered frames. This renders the coarse-SfM BA forward function as:
\begin{equation}
\resizebox{0.85\columnwidth}{!}{$
\begin{aligned}
\mathcal{L}_{\text{coarse}} &= -\frac{1}{N} \sum_i \frac{1}{\|R_i\|} \sum_{(j,k)\in \mathcal{G}_i}
\bar{F}_i(\bar{r}_{i,j,k}) \cdot \,\mathds{1}\!\left[\bar{r}_{i,j,k}<\bar{\tau}_{\text{max}}\right],
\end{aligned}
$}
\label{eqn:coarse_ba}
\end{equation}
where $\bar{r}_{i, j, k} = \log(1+r_{i, j, k})$ and $\bar{F}_i$ is the CDF  for the empirical distribution of logged residuals over the subgraph \( \mathcal{G}_i \).
Variable $R_i$ is the set of valid pixels included in subgraph \( \mathcal{G}_i \) up to a predefined maximum $\bar{\tau}_{\text{max}}$.

\Paragraph{Fine-Stage SfM.}
As shown in \cref{fig:pipeline}, we use L2 norm BA objective, with forward and backward functions defined in \cref{eqn:fine_forward} and \cref{eqn:fine_backward}, respectively. 
We evaluate over the entire graph. 
For both the coarse and fine stages, BA runs by applying gradient descent with a fixed number of iterations.

\Paragraph{Extension to Camera Re-localization.}
In \cref{fig:camreloc}, camera re-localization estimates the poses of unregistered query images given a set of registered map frames.
We extend to it with two changes. 
(1)
Intrinsic initialization is skipped as provided. 
The poses and depth adjustments are initialized using the map frame with the highest visibility.
(2) The pose graph includes both map and query frames.
Gradients for map frames are disabled.
We use groundtruth depth map of map frames if provided.
Otherwise, we estimate depth maps using MDE, \textit{e.g.}, the Wayspots map-free re-localization~\cite{Brachmann_2023_CVPR}.


\Paragraph{Extension to RANSAC.}
Using parallelized GPU implementation, we initialize $64$ minimal solutions choosing the maximizer according to \cref{eqn:binary_scoring_funciton_thresholds_summation}.
Details in Supp.Sec.~\ref{sec:exp_ransac}.

\section{Experiments} \label{sec:applications}
We benchmark our method's pose accuracy on two fundamental 3D vision tasks: Structure-from-Motion and camera re-localization.
For the SfM task we evaluate using small-scale ETH3D~\cite{Schops_2019_CVPR} IMC2021~\cite{bi2021method}, ScanNet~\cite{ScanNet_Dai_2017_CVPR}, and T\&T~\cite{knapitsch2017tanks} datasets.
For camera re-localization we evaluate using the 7-Scenes~\cite{shotton2013scene} and Wayspots~\cite{Brachmann_2023_CVPR} datasets.
Unless otherwise specified, we use the SoTA model DUSt3R~\cite{Wang_2024_CVPR} as the default monocular depth estimator, while also supporting other MDEs (see \cref{tab:scannet} and Supp. Tab.~\ref{tab:eth3d_plus}).
Following~\cite{Wang_2024_CVPR}, we feed DUSt3R with two identical images and extract the z-channel of the resulting point cloud as the depth map.
We use RoMa~\cite{edstedt2024roma} for dense correspondence.

\vspace{1ex}
\Paragraph{Implementation Details.}
We optimize coarse and fine stages using Adam~\cite{kingma2014adam} for total $50\text{k}$ iterations with a learning rate of 1e-3.
We sample $\kappa = 200$ pixels at each frame pair. 
We set maximum residual value to $\bar{\tau}_{\text{max}} = 10$ for the coarse BA objective, \cref{eqn:coarse_ba}, and $\tau_{\text{max}} = 20$ for the fine BA objective, \cref{eqn:binary_scoring_funciton}.
We include image pairs with  $\nu \geq 15\%$ of their pixels co-visible. 
Correspondences with confidence $\chi > 0.2$ are considered valid.
More in Supp.Sec.~\ref{sec:supp_details}.


\subsection{Structure-from-Motion Evaluations}

\begin{table*}[t!]
    \centering\footnotesize
    \resizebox{\linewidth}{!}{
    \begin{tabular}{l|cc|cc|cc|cc|cc|cc|cc}
    \toprule
    \multirow{2}{*}{\bf Scene} & \multicolumn{2}{c|}{COLMAP~\cite{Schonberger_2016_CVPR}}& \multicolumn{2}{c|}{ACE-Zero~\cite{brachmann2024acezero}}& \multicolumn{2}{c|}{FlowMap~\cite{smith25flowmap}}& \multicolumn{2}{c|}{VGG-SfM~\cite{wang2024vggsfm}}& \multicolumn{2}{c|}{DF-SfM~\cite{He_2024_CVPR}}& \multicolumn{2}{c|}{MASt3R-SfM~\cite{duisterhofmast3rsfm}}&\multicolumn{2}{c}{MBA (Ours)}\\
    & RRA & RTA & RRA & RTA & RRA & RTA & RRA & RTA & RRA & RTA & RRA & RTA & RRA & RTA \\
    \midrule
    courtyard & $56.3$ & $60.0$ & $4.0$ & $1.9$ & $7.5$ & $3.6$ & $50.5$ & $51.2$ & $80.7$ & $74.8$ & $89.8$ & $64.4$ & $94.7$ & $94.7$\\
    delivery area & $34.0$ & $28.1$ & $27.4$ & $1.9$ & $29.4$ & $23.8$ & $22.0$ & $19.6$ & $82.5$ & $82.0$ & $83.1$ & $81.8$ & $83.1$ & $83.0$ \\
    electro & $53.3$ & $48.5$ & $16.9$ & $7.9$ & $2.5$ & $1.2$ & $79.9$ & $58.6$ & $82.8$ & $81.2$ & $100.0$ & $95.5$ & $95.6$ & $78.2$\\
    facade & $92.2$ & $90.0$ & $74.5$ & $64.1$ & $15.7$ & $16.8$ & $57.5$ & $48.7$ & $80.9$ & $82.6$ & $74.3$ & $75.3$ & $100.0$ & $99.2$\\
    kicker & $87.3$ & $86.2$ & $26.2$ & $16.8$ & $1.5 $ & $1.5$ & $100.0$ & $97.8$ & $93.5$ & $91.0$ & $100.0$ & $100.0$ & $100.0$ & $98.9$ \\
    meadow & $0.9$ & $0.9$ & $3.8$ & $0.9$ & $3.8$ & $2.9$ & $100.0$ & $96.2$ & $56.2$  & $58.1$ & $58.1$ & $58.1$ & $100.0$ & $58.1$ \\
    office & $36.9$ & $32.3$ & $0.9$ & $0.0$ & $0.9$ & $1.5$ & $64.9$ & $42.1$ & $71.1$ & $54.5$ & $100.0$ & $98.5$ & $100.0$ & $86.2$ \\
    pipes & $30.8$ & $28.6$ & $9.9$ & $1.1$ & $6.6$ & $12.1$ & $100.0$ & $97.8$ & $72.5$ & $61.5$ & $100.0$ & $100.0$ & $100.0$ & $96.7$ \\
    playground & $17.2$ & $18.1$ & $3.8$ & $2.6$ & $2.6$ & $2.8$ & $37.3$ & $40.8$ & $70.5$ & $70.1$ & $100.0$ & $93.6$ & $94.7$ & $93.8$  \\
    relief & $16.8$ & $16.8$ & $16.8$ & $17.0$ & $6.9$ & $7.7$ & $59.6$ & $57.9$ & $32.9$ & $32.9$ & $34.2$ & $40.2$ & $100.0$ & $98.9$  \\
    relief 2 & $11.8$ & $11.8$ & $7.3$ & $5.6$ & $8.4$ & $2.8$ & $69.9$ & $70.3$ & $40.9$ & $39.1$ & $57.4$ & $76.1$ & $100.0$ & $98.9$\\
    terrace & $100.0$ & $100.0$ & $5.5$ & $2.0$ & $33.2$ & $24.1$ & $38.7$ & $29.6$ &  $100.0$ & $99.6$ & $100.0$ & $100.0$ & $100.0$ & $100.0$\\
    terrains & $100.0$ & $99.5$ & $15.8$ & $4.5$ & $12.3$ & $13.8$ & $70.4$ & $54.9$ & $100.0$ & $91.9$ & $58.2$ & $52.5$ & $100.0$ & $95.4$\\
    \midrule
    Average & $49.0$ & $47.8$ & $16.4$ & $9.7$ & $10.1$ & $8.8$ & $65.4$ & $58.9$ & $74.2$ & $70.7$ & $81.2$ & $79.7$ & $\mathbf{97.3}$ & $\mathbf{90.2}$ \\
    \bottomrule
    \end{tabular}
    }
    \vspace{0ex}
    \caption{\textbf{Structure-from-Motion} benchmark on ETH3D dataset~\cite{Schops_2019_CVPR} in metrics RRA ($@5^{\circ}$) and RTA ($@5^{\circ}$) following MASt3R-SfM evaluation protocol. 
    See performance with other MDE models at Supp.Tab.~\ref{tab:eth3d_plus}.
    See detailed evaluation protocol at Supp.Sec.~\ref{sec:protocol}
    }
    \label{tab:eth3d}
\end{table*}

\begin{table}
\centering
\resizebox{\linewidth}{!}{%
\begin{tabular}{c|cccc}
    \toprule
    Type & Method & AUC$@3^\circ$ & AUC$@5^\circ$ & AUC$@10^\circ$ \\
    \midrule
    \multirow{5}{*}{Detector-Based} & SIFT+NN + COLMAP~\cite{Schonberger_2016_CVPR} \tiny{CVPR'16} & $24.87$ & $34.47$ & $45.94$ \\
    & SIFT + NN + PixSfM~\cite{lindenberger2021pixsfm} \tiny{ICCV'21} & $26.45$ & $35.73$ & $47.24$ \\
    & D2Net + NN + PixSfM~\cite{lindenberger2021pixsfm}\tiny{ICCV'21} & $10.27$ & $13.12$ & $17.25$ \\
    & R2D2 + NN + PixSfM~\cite{lindenberger2021pixsfm} \tiny{ICCV'21}& $32.44$ & $42.55$ & $55.01$ \\
    & SP + SG + PixSfM~\cite{lindenberger2021pixsfm} \tiny{ICCV'21}& $46.30$ & $58.43$ & $71.62$ \\
    \midrule
    \multirow{4}{*}{Detector-Free} & LoFTR + PixSfM~\cite{lindenberger2021pixsfm} \tiny{ICCV'21}& $44.06$ & $56.16$ & $69.61$ \\
    & LoFTR + DF-SfM~\cite{He_2024_CVPR} \tiny{CVPR'24} & $46.55$ & $58.74$ & $72.19$ \\
    & AspanTrans. + DF-SfM~\cite{He_2024_CVPR}\tiny{CVPR'24} & $46.79$ & $59.01$ & $72.50$ \\
    & MatchFormer + DF-SfM~\cite{He_2024_CVPR}\tiny{CVPR'24} & $45.83$ & $57.88$ & $71.22$ \\
    \midrule
    \multirow{2}{*}{Dense Matching} & DKM + Dense-SfM~\cite{lee2025densesfmstructuremotiondense}\tiny{CVPR'25} & $\underline{48.65}$ & $\underline{61.09}$ & ${74.37}$ \\
    & RoMa + Dense-SfM~\cite{lee2025densesfmstructuremotiondense}\tiny{CVPR'25} & ${48.48}$ & ${60.79}$ & ${73.90}$ \\
    \midrule
    \multirow{3}{*}{Deep-based} & VGG-SfM~\cite{wang2024vggsfm}\tiny{CVPR'24} & $45.23$ & $58.89$ & ${73.92}$ \\
    & VGGT~\cite{wang2025vggt} \tiny{CVPR'25} & $39.23$ & $52.74$ & $71.26$ \\
    & VGGT~\cite{wang2025vggt}~+~BA \tiny{CVPR'25} & $\mathbf{66.37}$ & $\mathbf{75.16}$ & $\mathbf{84.91}$ \\
    \midrule
    \multirow{2}{*}{Point-Based} & Mast3r-SfM~\cite{duisterhofmast3rsfm}\tiny{3DV'25} & $42.26$ & $54.53$ & $67.97$ \\
    & Dense-SfM + Mast3r-SfM~\cite{duisterhofmast3rsfm}\tiny{3DV'25} & $44.98$ & $57.09$ & $70.09$ \\
    \midrule
    MDE & MBA (Ours) & $47.29$ & $60.37$ & $\underline{74.59}$ \\
    \bottomrule
\end{tabular}
}
\caption{\textbf{Structure-from-Motion} on IMC2021~\cite{bi2021method} dataset in AUC metric.
We are comparable or better than Dense-SfM, DF-SfM, Mast3r-SfM, PixSfM, VGG-SfM and VGGT without BA.
\vspace{-2mm}}
\label{tab:imc}
\end{table}

\Paragraph{Benchmark on ETH3D.}
ETH3D~\cite{Schops_2019_CVPR} contains unordered, high-resolution indoor and outdoor images with precisely calibrated groundtruth camera poses. 
In \Cref{tab:eth3d} we follow the protocol of MASt3R-SfM~\cite{duisterhofmast3rsfm} in reporting the relative rotation accuracy (RRA) and relative translation accuracy (RTA) of poses at a fixed threshold. 
We achieve SoTA with substantial improvement in \cref{tab:eth3d}.
We outperform the classic  COLMAP~\cite{Schonberger_2016_CVPR}, DF-SfM~\cite{He_2024_CVPR}, and the scene-coordinate regression based method ACE-Zero~\cite{brachmann2024acezero}. 
We also outperform the learning-based VGG-SfM~\cite{wang2024vggsfm}, and its counterparts FlowMap~\cite{smith25flowmap} and MASt3R-SfM~\cite{duisterhofmast3rsfm}, using depth maps and point clouds respectively.
Notably, we outperform MASt3R-SfM using the same depth estimator of DUSt3R and a less-performant RoMa correspondence estimator.
Our experiments indicate that the zero-shot monocular depth map already produces comparable or better performance than classic SfM algorithms.

\Paragraph{Benchmark on IMC2021.}
The IMC2021~\cite{bi2021method} dataset features internet images of tourist landmarks, organized into $1,020$ subsets each containing $5$, $10$, or $25$ images. 
\cref{fig:teaser} shows example results of London Bridge. 
Despite including challenging elements including sky, river, tourists and trees, MDE depth maps still enable competitive pose performance.
\Cref{tab:imc} presents quantitative results on IMC2021. 
Our method performs second to VGGT~\cite{wang2025vggt}~+~BA at AUC$@10^\circ$ and competitive results at other thresholds.
We outperform the learning based methods VGG-SfM~\cite{wang2024vggsfm}, and FlowMap~\cite{smith25flowmap} while consistently outperforming MASt3R-SfM~\cite{duisterhofmast3rsfm}.
Our method also ranks similarly with SoTA classic method Dense-SfM~\cite{lee2025densesfmstructuremotiondense} while outperform others including COLMAP~\cite{Schonberger_2016_CVPR}, PixSfM~\cite{lindenberger2021pixsfm}, and DF-SfM~\cite{He_2024_CVPR}.
We observe that VGGT+BA significantly improves performance, suggesting BA remains a necessary component for foundation feed-forward SfM models. 
We contribute a new BA objective that exploits the density of per-pixel regressions.
It potentially applies to recent feed-forward systems including VGG-SfM, VGGT
, and Light3R-SfM~\cite{elflein2025light3r} that already output dense depth maps or point clouds.
Finally, we scale to large optimization problem (\cref{tab:scalability}) while feed-forward baselines FlowMap and VGG-SfM run out of memory with more than $200$ views~\cite{duisterhofmast3rsfm}.

\begin{table}[!tp]
\begin{center}
\resizebox{\linewidth}{!}{
\sisetup{detect-all=true,detect-weight=true}
\begin{tabular}{l|c|cccc}
\toprule
 Method  & Inference & RRA@5 $\uparrow$ & RTA@5 $\uparrow$ & ATE $\downarrow$ & Registration $\uparrow$ \\
\midrule
Cut3R~\cite{wang2025continuous}~\tiny{CVPR'25}          & \multirow{6}{*}{\rotatebox{90}{FeedForward}} & $18.8$ & $25.8$ & $0.017$ &  $100.0$ \\
Spann3R~\cite{wang20253d}~\tiny{3DV'25}                &                                                & $22.1$ & $30.7$ & $0.016$ &  $100.0$ \\
SLAM3R~\cite{liu2025slam3r}~\tiny{CVPR'25}             &                                                & $20.3$ & $24.7$ &  $0.015$ &  $100.0$ \\
VGGT-SLAM~\cite{maggio2025vggt}~\tiny{arXiv'25}        &                                                & $57.3$ & $67.9$ &  $\mathbf{0.008}$ &  $100.0$ \\
Light3R-SfM~\cite{elflein2025light3r}~\tiny{CVPR'25}   &                                                & $52.0$ & $52.8$ &  $0.011$ &  $100.0$ \\
SAIL~\cite{deng2025sail}~\tiny{arXiv'25}               &                                                & $70.4$ & $74.7$ &  $\mathbf{0.008}$ &  $100.0$ \\
\midrule
GLOMAP~\cite{pan2024glomap}~\tiny{ECCV'24}             & \multirow{9}{*}{\rotatebox{90}{Optimization}}  & $75.8$ & $76.7$ & $0.010$ & $100.0$ \\
ACE0~\cite{brachmann2024acezero}~\tiny{ECCV'24}        &                                                & $56.9$ & $57.9$ & $0.015$ & $100.0$ \\
DF-SfM~\cite{He_2024_CVPR}~\tiny{CVPR'24}              &                                                & $69.6$ & $69.3$ & $0.014$ & $76.2$ \\
FlowMap~\cite{smith25flowmap}~\tiny{3DV'25}            &                                                & $31.7$ & $35.7$ & $0.017$ & $66.7$ \\
VGGSfM~\cite{wang2024vggsfm}~\tiny{CVPR'24}            &                                                & $-$    & $-$    & $-$     & $0.0$ \\
MASt3R-SfM~\cite{duisterhofmast3rsfm}~\tiny{3DV'25}    &                                                & $49.2$ & $54.0$ & $0.011$ & $100.0$ \\
DROID-SLAM~\cite{teed2021droid}~\tiny{NeurIPS'21}      &                                                & $31.3$ & $40.3$ & $0.021$ & $100.0$ \\
SAIL-OPT~\cite{deng2025sail}~\tiny{arXiv'25}           &                                                & $71.5$ & $\mathbf{77.7}$ & $\mathbf{0.008}$ & $100.0$ \\
MBA (ours)                                             &                                                & $\mathbf{71.7}$ & $77.0$ & $0.009$ & $100.0$ \\
\bottomrule
\end{tabular}
}
\caption{\textbf{Structure-from-Motion} on Tanks\&Temple~\citep{knapitsch2017tanks} dataset.}
\label{tab:tnt_pose}
\end{center}
\vspace{-4mm}
\end{table}

\begin{table}[t!]
    \centering\small
    \resizebox{\linewidth}{!}{%
    \begin{tabular}{l|l|l|ccc}
    \toprule
    {\bf Method} &Depth&Corres.&$\text{ACC}@3^\circ$&$\text{ACC}@5^\circ$&$\text{ACC}@10^\circ$\\ 
    \midrule
    COLMAP~\cite{Schonberger_2016_CVPR}&- & SuperPoint~\cite{detone2018superpoint} & ${0.342}$ & ${0.505}$ & $0.670$ \\
    \midrule
    \multirow{5}{*}{MBA (Ours)}
    & ZoeDepth~\cite{bhat2023zoedepth} & RoMa~\cite{edstedt2024roma} & ${0.372}$ & ${0.586}$ & ${0.811}$ \\
    & DUSt3R~\cite{Wang_2024_CVPR}  & RoMa~\cite{edstedt2024roma} & $0.403$ & $\mathbf{0.615}$ & $0.820$ \\
    & UniDepth~\cite{piccinelli2024unidepth} & RoMa~\cite{edstedt2024roma} & $\mathbf{0.407}$ & $0.612$ & $\mathbf{0.823}$ \\
    & DUSt3R~\cite{Wang_2024_CVPR}  & MASt3R~\cite{mast3r_arxiv24} & $0.384$ & $0.596$ & $0.811$ \\
    & UniDepth~\cite{piccinelli2024unidepth} & MASt3R~\cite{mast3r_arxiv24} & $0.393$ & $0.598$ & $0.817$ \\
    \bottomrule
    \end{tabular}
    }
    \caption{
    \textbf{Structure-from-Motion} on the ScanNet dataset~\cite{ScanNet_Dai_2017_CVPR}. 
    }
    \vspace{-4mm}
    \label{tab:scannet}
\end{table}

\Paragraph{Benchmark on Tanks\&Temples.}
T\&T~\cite{knapitsch2017tanks} is a large-scale SfM benchmark including $21$ scenes, of around $300$ images each, with pseudo-groundtruth from COLMAP. 
We follow concurrent work SAIL~\cite{deng2025sail} in evaluation, where VGGSfM fails to converge on this benchmark.
Our MDE-anchored method performs better or on-par with both feedforward and optimization baselines; see \cref{tab:tnt_pose}.

\Paragraph{Benchmark on ScanNet.}
We further evaluate different combinations of MDE and correspondence models on 
ScanNet at \cref{tab:scannet}.
We compare with COLMAP only on frames it successfully registers, and our MDE-based SfM system still outperforms it on this large-scale SfM benchmark.


\Paragraph{Benchmark on FastMap.}
We additionally compare to SoTA classic SfMs~\cite{fastmap2025, pan2024glomap} on large scenes in Supp.\cref{tab:fastmap_comparison}. 

\begin{table}[t!]
    \centering
    \small
    \resizebox{0.7\linewidth}{!}{%
    \begin{tabular}{ll|c}
    \toprule
    {\bf Category} & {\bf Method} & Average (deg/cm)\\
    \midrule
    \multirow{2}{*}{FM} & AS~\cite{sattler2016efficient}\tiny{PAMI'16} & $5.1/2.46$\\
     &HLoc~\cite{sarlin2019coarse}\tiny{CVPR'19} & $\mathbf{3.4}/\mathbf{1.07}$ \\
    \midrule
    \multirow{3}{*}{E2E}
    & SC-wLS~\cite{wu2022sc}\tiny{ECCV'22} & $6.6/1.45$ \\
    & NeuMaps~\cite{Tang_2023_CVPR}\tiny{CVPR'23} & $3.1/1.09$ \\
    & PixLoc~\cite{Sarlin_2021_CVPR}\tiny{CVPR'21} & $\mathbf{2.9}/\mathbf{0.98}$\\
    \midrule
    \multirow{4}{*}{SCR}
    & ACE~\cite{Brachmann_2023_CVPR}\tiny{CVPR'23} & $2.8/0.93$ \\
    & DSAC*~\cite{DSAC_PAMI2022}\tiny{PAMI'22} & $2.7/1.41$\\
    & HSCNet~\cite{Li_2020_CVPR}\tiny{CVPR'20} & $2.7/0.90$\\
    & HSCNet++~\cite{Wang2024}\tiny{IJCV'24} & $\mathbf{2.29}/\mathbf{0.81}$\\
    \midrule
    \multirow{1}{*}{APR}
    & MAREPO~\cite{marepo_Chen_2024_CVPR}\tiny{CVPR'24} & $\mathbf{3.2}/\mathbf{1.54}$ \\
    \midrule
    MDE & MBA (Ours) & $\mathbf{2.7}/\mathbf{0.92}$ \\
    \bottomrule
    \end{tabular}
    }
    \caption{\textbf{Camera relocalization} performance benchmark on the 7-Scenes dataset~\cite{shotton2013scene}. Per-scene results are in Supp.Tab.~\ref{tab:seven_scenes_full}.
    \vspace{-4mm}}
    \label{tab:seven_scenes}
\end{table}


\subsection{Camera Re-Localization Evaluations}

\Paragraph{Benchmark on 7-Scenes.}
For the task of camera relocalization on 7-Scenes, we follow DUSt3R and MAREPO~\cite{marepo_Chen_2024_CVPR} in categorizing baselines into four groups: feature matching (FM), end-to-end (E2E), scene coordinate regression (SCR), and absolute pose regression (APR). 
We follow DUSt3R in using DIR~\cite{gordo2016deep} retrieving the top $200$ query-to-query and query-to-map images in evaluating our method.
\Cref{tab:seven_scenes} presents relocalization metrics, ranking our method second to HSCNet++~\cite{Wang2024}, while outperforming
ACE~\cite{Brachmann_2023_CVPR}, DSAC*~\cite{DSAC_PAMI2022}, MAREPO~\cite{marepo_Chen_2024_CVPR}, and HLoc~\cite{sarlin2019coarse}. 
Note that our method is scene-agnostic, whereas HSCNet++ is a scene-specific approach. 
We note too that our method is a multi-view approach that assumes multiple test images are available, while baselines operate on each query independently.

\Paragraph{Benchmark on Wayspots.}
Wayspots~\cite{Brachmann_2023_CVPR} is a map-free camera relocalization dataset comprised of handheld smartphones images.
Unlike 7-Scenes, it does not provide groundtruth depth map for map images.
We estimate depth maps for query and map images with DUSt3R and construct a pose graph via exhaustive image matching. 
\Cref{tab:wayspots} presents our results on Wayspots.
Compared to baselines, we directly use rotated images without alignment (
See example at \cref{fig:camreloc}), 
yet still achieve SoTA performance. 
This highlights our method’s ability to leverage the generalization power of the MDE model.
Baselines may face challenge as the scene-specific finetuning is nevertheless constrained to the texture and lighting conditions of the video.

\begin{table}[t!]
    \centering
    \small
\resizebox{\linewidth}{!}{%
\begin{tabular}{l|cccc}
    \toprule
    Scene & MAREPO~\cite{marepo_Chen_2024_CVPR} & $\text{DSAC}^{*}$~\cite{DSAC_PAMI2022} & ACE~\cite{Brachmann_2023_CVPR} & MBA (Ours)  \\
    \midrule
    Cubes & $71.8\%$ &  $83.8\%$ & $\mathbf{97.0\%}$ & $75.1\%$ \\
    Bears & $80.7\%$ & $82.6\%$ & $80.7\%$ & $\mathbf{100\%}$ \\
    Winter Sign & $0.0\%$ & $0.2\%$ & $1.0\%$ & $\mathbf{9.3\%}$ \\
    Inscription & $37.1\%$ & $54.1\%$ & $\mathbf{49.0\%}$ & $28.3\%$ \\
    The Rock & $99.8\%$ & $\mathbf{100\%}$ & $\mathbf{100\%}$ & $\mathbf{100\%}$\\
    Tendrils & $29.3\%$ & $25.1\%$ &  $34.9\%$ & $\mathbf{51.5\%}$ \\
    Map & $55.1\%$ & $56.7\%$ & $\mathbf{56.5\%}$ & $45.1\%$\\
    Square Bench & $\mathbf{70.7\%}$ & $69.5\%$ & ${66.7\%}$ & ${58.6\%}$ \\
    Statue & $0.0\%$ & $0.0\%$ & $0.0\%$ & $0.0\%$ \\
    Lawn & $34.2\%$ & $34.7\%$ & $35.8\%$ & $\mathbf{85.0\%}$\\
    \midrule
    Average-Accuracy & $47.9\%$ & $50.7\%$ & $52.2\%$ & $\mathbf{55.29\%}$\\
    \bottomrule
\end{tabular}
}
\caption{\textbf{Camera relocalization} benchmark on the Wayspots~\cite{Brachmann_2023_CVPR}.
\vspace{-2mm}
}
\label{tab:wayspots}
\end{table}

\begin{table}
\small
    \centering
    \resizebox{\linewidth}{!}{%
    \begin{tabular}{l|ccc|ccc}
    \toprule
    & \multicolumn{3}{c|}{\textbf{MegaDepth}} & \multicolumn{3}{c}{\textbf{ScanNet}} \\
    Method $\downarrow$\quad\quad\quad AUC$@$ $\rightarrow$
    &$5^{\circ}$ $\uparrow$&$10^{\circ}$ $\uparrow$&$20^{\circ}$ $\uparrow$
    &$5^{\circ}$ $\uparrow$&$10^{\circ}$ $\uparrow$&$20^{\circ}$ $\uparrow$\\
    \midrule         
    LoFTR~\citep{sun2021loftr}~\tiny{CVPR'21}& 52.8 & 69.2 & 81.2 & 22.1 & 40.8 & 57.6\\
    PDC-Net+~\citep{truong2023pdc}~\tiny{TPAMI'23} & 51.5 & 67.2 & 78.5 & 20.3 & 39.4 & 57.1\\
    DKM~\citep{edstedt2023dkm}~\tiny{CVPR'23} & 60.4 & 74.9 & 85.1 & 29.4 & 50.7 & 68.3\\
    PMatch~\citep{zhu2023pmatch}~\tiny{CVPR'23} & 61.4 & 75.7 & 85.7 & 29.4 & 50.1 & 67.4\\
    RoMa~\cite{edstedt2024roma}~\tiny{CVPR'24} & {62.6} & {76.7} & {86.3} & {31.8} & {53.4} & {70.9}\\
    \hline
    RoMa~\cite{edstedt2024roma} + MAGSAC++~\cite{barath2020magsac++} & $\mathbf{68.0}$ & $\mathbf{79.8}$ & $\mathbf{88.0}$ & $\mathbf{32.9}$ & $\mathbf{54.6}$ & $\mathbf{71.3}$\\
    RoMa~\cite{edstedt2024roma} + MBA (Ours) & $\underline{66.9}$ & $\underline{79.3}$ & $\underline{87.6}$ & $\underline{32.4}$ & $\underline{54.1}$ & $\underline{71.0}$\\
    \bottomrule
    \end{tabular}
    }
    \caption{\textbf{RoMa benchmark} of two-view pose estimation with RANSAC on MegaDepth-1500~\citep{li2018megadepth} and ScanNet-1500~\citep{dai2017scannet}.
    \vspace{-4mm}} 
    \label{tab:ransac_combined}
\end{table}


\subsection{Ablation Study}
\Paragraph{MDE Model.}
Our method accomodates different MDEs, and performance consistently improves with stronger ones. 
On ScanNet (\cref{tab:scannet}), replacing ZoeDepth with UniDepth increases Acc$@3^\circ$ from $0.396$ to $0.439$.
A similar improvement is observed on ETH3D; see Supp.Tab.~\ref{tab:eth3d_plus}.

\Paragraph{Two-View RANSAC.}
Shown in the \cref{tab:ransac_combined}, the RANSAC-inspired scoring function \cref{eqn:binary_scoring_funciton_thresholds} serves as an alternative scoring function in RANSAC, showing a comparable performance with the SoTA method MAGSAC++~\cite{barath2020magsac++}.

\Paragraph{Algorithmic Choices.}
In Supp.Tab.~\ref{tab:optimizer}, we compare different optimization schemes and further demonstrate that increasing the sampling density yields marginal improvements in pose estimation accuracy. Finally, we compare the MBA loss to conventional robust loss functions~\cite{schoenberger2016sfm, mlotshwa2022cauchy} where MBA loss consistently outperforms others.
\section{Conclusion} 
\label{sec:conclusions}

We introduced ``Marginalized Bundle Adjustment'' (MBA), a multi-view pose estimation method that leverages monocular depth estimators (MDE).
Our core contribution is a RANSAC-motivated BA objective enabled by dense network predictions.
Our method scales to large datasets and generalizes across scenes, highlighting the potential of MDE in multi-view pose estimation task.

\Paragraph{Limitations.}
Our use of a first-order optimizer leads to higher runtime compared to other SfM baselines.
Beyond depth estimation models, it would be interesting to investigate a tight integration of our proposed MBA with feed-forward foundation models such as VGGT.

\vspace{-9ex}
{
    \small
    \bibliographystyle{ieeenat_fullname}
    \bibliography{main}
}

\clearpage
\setcounter{page}{1}
\maketitlesupplementary

\begin{table}[!t]
    \centering
    \resizebox{0.85\linewidth}{!}{%
    \begin{tabular}{l|cc|cc}
    \toprule
    \multirow{2}{*}{Dataset} & \multicolumn{2}{c|}{Max Graph Size} & \multicolumn{2}{c}{Max Runtime} \\
    & Images & Pairs & Ours & COLMAP \\
    \midrule
    ScanNet~\cite{ScanNet_Dai_2017_CVPR} & $391$ & $43{,}964$ & $2.98$h & $0.91$h \\
    ETH3D~\cite{Schops_2019_CVPR} & $76$ & $3{,}142$ & $41$min & $9$min \\
    IMC2021~\cite{bi2021method} & $25$ & $600$ & $14$min & $3$min \\
    7-Scenes~\cite{shotton2013scene} & $8{,}000$ & $564{,}418$ & $3.12$h & $13.52$h \\
    Wayspots~\cite{Brachmann_2023_CVPR} & $1{,}157$ & $1{,}333{,}196$ & $4.72$h & Crash \\
    \bottomrule
    \end{tabular}
    }
    \caption{
    \textbf{Max Pose Graph and its Runtime} in each benchmarked dataset.
    We run experiments on ScanNet, ETH3D, and IMC2021 with single $\text{V}100$ GPU, and on 7-Scenes and Wayspots with eight.
    }
    \label{tab:scalability_supp}
    \vspace{-1ex}
\end{table}

\section{Extended Methodology} \label{sec:extended_methodology}

\Paragraph{Compare Surrogate Loss to MAGSAC.}
\label{sec:surrogate_loss_to_magsac}
We rewrite the surrogate loss \cref{eqn:fine_forward} into an integration form:
\begin{equation}
\begin{aligned}
    \mathcal{L}_{\text{MBA}}
    &= \frac{1}{\|R\|} \sum_{i, j, k} -F(r_{i, j, k}) \cdot \mathds{1}[r_{i, j, k} < \tau_{\text{max}}] \\
    &= - \int_{0}^{\tau_{\text{max}}} F(r) \cdot p(r) \,\, d r .
\end{aligned}
\label{eqn:fine_forward_magsac}
\end{equation}
The formulation in \cref{eqn:fine_forward_magsac} resembles the scoring function presented by MAGSAC~\cite{barath2019magsac}.
In \cref{eqn:fine_forward_magsac}, we marginalize the residual probability density function $p(r)$ with its empirical CDF function $F(r)$.
MAGSAC~\cite{barath2019magsac} instead replaces the empirical CDF function $F(r)$ with a fixed chi-square distribution derived from assumptions on the distribution of residual errors for inliers and outliers. 
Despite their similarity, \cref{eqn:fine_forward} only serves as a surrogate forward loss for our BA objective.
We compare RANSAC performance using the proposed scoring function \cref{eqn:binary_scoring_funciton_thresholds} against MAGSAC in \cref{tab:ransac_combined}.
Our more generalized scoring function demonstrates comparable performance. 
Please refer to \cref{sec:exp_ransac} for details of the experiments.


\section{Extended Experiments} \label{sec:additional_experiments}

\subsection{Two-View RANSAC}
\label{sec:exp_ransac}

We follow the RoMa~\cite{edstedt2024roma} evaluation protocol to benchmark RANSAC-based essential matrix estimation. 
Results on the MegaDepth-1500~\cite{li2018megadepth} and ScanNet-1500~\cite{dai2017scannet} benchmarks are reported in \cref{tab:ransac_combined}. 
Replacing RoMa’s default RANSAC, we evaluate MAGSAC++ and our method under identical budgets, tuning both over the same inlier-threshold grid and sampling the same number of correspondences. 
For both methods, we sample $1\,000$ correspondences and sweep the inlier threshold over the grid $\{0.02,\ 0.05,\ 0.1,\ 0.2,\ 0.3,\ 0.5,\ 0.8,\ 1.0,\ 1.2,\ 1.5,\ 1.7,\ 2.0\}$ in normalized pixel coordinates. 
Our method uses a GPU-parallelized estimator: we always randomly compute $64$ minimal solutions in parallel and retain the one that maximizes the scoring function in \cref{eqn:binary_scoring_funciton_thresholds_summation}. 
We use OpenCV's MAGSAC++ implementation.
Both MAGSAC++ and our method substantially outperform standard RANSAC, and our alternative scoring function~\cref{eqn:binary_scoring_funciton_thresholds_summation} achieves performance on par with MAGSAC++. 
These results indicate that our RANSAC-motivated scoring function applies to two-view essential matrix estimation, beyond the multi-view pose setting.

\begin{table}[!t]
    \centering
    \small
\resizebox{0.75\linewidth}{!}{%
\begin{tabular}{c|l|cc}
    \toprule
    Ablation & Method & RRA$@5^\circ$ & RTA$@5^\circ$ \\
    \midrule
    \multirow{3}{*}{\rotatebox{90}{{Stages}}} 
    & Initialization & $62.6$ & $41.0$ \\
    & Coarse & $97.1$ & $87.9$ \\
    & Fine & $97.3$ & $90.2$ \\
    \midrule
    \multirow{3}{*}{\rotatebox{90}{{Density}}} 
    & $\kappa = 30$ & $91.7$ & $78.6$ \\
    & $\kappa = 200$ & $97.3$ & $90.2$ \\
    & $\kappa = 500$ & $99.3$ & $92.0$ \\
    \midrule
    \multirow{5}{*}{\rotatebox{90}{{Losses}}} 
    & Soft L1 & $87.5$ & $73.9$ \\
    & Cauchy & $86.2$ & $75.2$ \\
    & Tukey & $63.5$ & $40.7$ \\
    & L2 & $79.4$ & $66.0$ \\
    & MBA (ours) & $97.3$ & $90.2$ \\
    \bottomrule
\end{tabular}
}
\vspace{-2ex}
\caption{
\footnotesize
\textbf{Ablations} on ETH3D dataset.
}
\vspace{-2ex}
\label{tab:optimizer}
\end{table}

\begin{table}
\centering
\resizebox{\linewidth}{!}{%
\begin{tabular}{c|cccc}
    \toprule
    \multirow{2}{*}{Type} & \multirow{2}{*}{Method} & \multicolumn{3}{c}{ETH3D Dataset} \\
    \cline{3-5}
    & & AUC$@1^\circ$ & AUC$@3^\circ$ & AUC$@5^\circ$ \\
    \midrule
    \multirow{5}{*}{Detector-Based} & SIFT+NN + COLMAP~\cite{Schonberger_2016_CVPR} \tiny{CVPR'16} & $26.71$ & $38.86$ & $42.14$ \\
    & SIFT + NN + PixSfM~\cite{lindenberger2021pixsfm} \tiny{ICCV'21} & $26.94$ & $39.01$ & $42.19$ \\
    & D2Net + NN + PixSfM~\cite{lindenberger2021pixsfm}\tiny{ICCV'21} & $34.50$ & $49.77$ & $53.58$ \\
    & R2D2 + NN + PixSfM~\cite{lindenberger2021pixsfm} \tiny{ICCV'21}& $43.58$ & $62.09$ & $66.89$ \\
    & SP + SG + PixSfM~\cite{lindenberger2021pixsfm} \tiny{ICCV'21}& $50.82$ & $68.52$ & $72.86$ \\
    \midrule
    \multirow{4}{*}{Detector-Free} & LoFTR + PixSfM~\cite{lindenberger2021pixsfm} \tiny{ICCV'21}& $54.35$ & $73.97$ & $78.86$ \\
    & LoFTR + DF-SfM~\cite{He_2024_CVPR} \tiny{CVPR'24} & $59.12$ & $75.59$ & $79.53$ \\
    & AspanTrans. + DF-SfM~\cite{He_2024_CVPR}\tiny{CVPR'24} & $57.23$ & $73.71$ & $77.70$ \\
    & MatchFormer + DF-SfM~\cite{He_2024_CVPR}\tiny{CVPR'24} & $56.70$ & $73.00$ & $76.84$ \\
    \midrule
    \multirow{2}{*}{Dense Matching} & DKM + Dense-SfM~\cite{lee2025densesfmstructuremotiondense}\tiny{CVPR'25} & $59.04$ & $77.73$ & $82.20$ \\
    & RoMa + Dense-SfM~\cite{lee2025densesfmstructuremotiondense}\tiny{CVPR'25} & $\mathbf{60.92}$ & $\mathbf{78.41}$ & $\mathbf{82.63}$ \\
    \midrule
    Deep-based & VGG-SfM~\cite{wang2024vggsfm}\tiny{CVPR'24} & \multicolumn{3}{c}{(compared in \cref{tab:eth3d})} \\
    \midrule
    \multirow{2}{*}{Point-Based} & Mast3r-SfM~\cite{duisterhofmast3rsfm}\tiny{arXiv'24} & $35.85$ & $58.46$ & $65.03$ \\
    & Dense-SfM + Mast3r-SfM~\cite{duisterhofmast3rsfm}\tiny{arXiv'24}  & $37.50$ & $59.18$ & $65.48$ \\
    \midrule
    MDE & MBA (Ours) & $27.72$ & $63.35$ & $74.02$ \\
    \bottomrule
\end{tabular}
}
\caption{\textbf{Structure-from-Motion} on ETH3D~\cite{Schops_2017_CVPR, Schops_2019_CVPR} dataset in metric AUC at multiple thresholds following DF-SfM.}
\label{tab:eth3d_auc}
\end{table}

\begin{table*}[t]
\resizebox{\linewidth}{!}{
\tablestyle{2pt}{1.1}
\begin{tabular}{r r | r@{~}r@{~}r@{~}r | r@{~}r@{~}r@{~}r | r@{~}r@{~}r@{~}r | r@{~}r@{~}r@{~}r | r@{~}r@{~}r@{~}r}

\toprule

  &  &
\multicolumn{4}{c}{ATE$\downarrow$} &
\multicolumn{4}{c}{RTA@3$\uparrow$} &
\multicolumn{4}{c}{AUC-R\&T @ 3 $\uparrow$} &
\multicolumn{4}{c}{RTA@1$\uparrow$} &
\multicolumn{4}{c}{AUC-R\&T @ 1 $\uparrow$} \\

\cmidrule(lr){3-6} \cmidrule(lr){7-10} \cmidrule(lr){11-14}
\cmidrule(lr){15-18} \cmidrule(lr){19-22}

  & n\_imgs &
\multicolumn{1}{c}{\scriptsize MBA (Ours)} & \multicolumn{1}{c}{\scriptsize \alg~\cite{fastmap2025}} & \multicolumn{1}{c}{\scriptsize GLOMAP~\cite{pan2024glomap}} & \multicolumn{1}{c}{\scriptsize COLMAP~\cite{Schonberger_2016_CVPR}} &
\multicolumn{1}{c}{\scriptsize MBA (Ours)} & \multicolumn{1}{c}{\scriptsize \alg~\cite{fastmap2025}} & \multicolumn{1}{c}{\scriptsize GLOMAP~\cite{pan2024glomap}} & \multicolumn{1}{c}{\scriptsize COLMAP~\cite{Schonberger_2016_CVPR}} &
\multicolumn{1}{c}{\scriptsize MBA (Ours)} & \multicolumn{1}{c}{\scriptsize \alg~\cite{fastmap2025}} & \multicolumn{1}{c}{\scriptsize GLOMAP~\cite{pan2024glomap}} & \multicolumn{1}{c}{\scriptsize COLMAP~\cite{Schonberger_2016_CVPR}} &
\multicolumn{1}{c}{\scriptsize MBA (Ours)} & \multicolumn{1}{c}{\scriptsize \alg~\cite{fastmap2025}} & \multicolumn{1}{c}{\scriptsize GLOMAP~\cite{pan2024glomap}} & \multicolumn{1}{c}{\scriptsize COLMAP~\cite{Schonberger_2016_CVPR}} &
\multicolumn{1}{c}{\scriptsize MBA (Ours)} & \multicolumn{1}{c}{\scriptsize \alg~\cite{fastmap2025}} & \multicolumn{1}{c}{\scriptsize GLOMAP~\cite{pan2024glomap}} & \multicolumn{1}{c}{\scriptsize COLMAP~\cite{Schonberger_2016_CVPR}} \\

\midrule
     mipnerf360  (9) & 215.6 &  \cellCD[22pt]{ffffff}{5.0e-4} &  \cellCD[22pt]{ffffff}{4.2e-4} &  \cellCD[22pt]{69a84f}{3.3e-5} &  \cellCD[22pt]{b7d7a8}{5.8e-5} &  \cellCD[22pt]{b7d7a8}{99.6} &  \cellCD[22pt]{b7d7a8}{99.9} &  \cellCD[22pt]{69a84f}{100.0} &  \cellCD[22pt]{69a84f}{100.0} &  \cellCD[22pt]{ffffff}{85.5} &  \cellCD[22pt]{b7d7a8}{97.4} &  \cellCD[22pt]{69a84f}{98.2} &  \cellCD[22pt]{b7d7a8}{97.2} &  \cellCD[22pt]{ffffff}{87.3} &  \cellCD[22pt]{b7d7a8}{99.8} &  \cellCD[22pt]{69a84f}{100.0} &  \cellCD[22pt]{b7d7a8}{99.7} &  \cellCD[22pt]{ffffff}{66.1} &  \cellCD[22pt]{b7d7a8}{92.4} &  \cellCD[22pt]{69a84f}{94.6} &  \cellCD[22pt]{b7d7a8}{91.9} \\
  tnt\_advanced  (6) & 337.8 &  \cellCD[22pt]{ffffff}{1.7e-2} &  \cellCD[22pt]{b7d7a8}{6.4e-3} &  \cellCD[22pt]{b7d7a8}{1.2e-2} &  \cellCD[22pt]{69a84f}{1.2e-3} &  \cellCD[22pt]{ffffff}{59.9} &  \cellCD[22pt]{ffffff}{71.4} &  \cellCD[22pt]{b7d7a8}{79.1} &  \cellCD[22pt]{69a84f}{98.5} &  \cellCD[22pt]{ffffff}{34.7} &  \cellCD[22pt]{ffffff}{42.6} &  \cellCD[22pt]{b7d7a8}{75.3} &  \cellCD[22pt]{69a84f}{94.8} &  \cellCD[22pt]{ffffff}{25.2} &  \cellCD[22pt]{ffffff}{42.3} &  \cellCD[22pt]{b7d7a8}{77.5} &  \cellCD[22pt]{69a84f}{97.0} &  \cellCD[22pt]{ffffff}{12.5} &  \cellCD[22pt]{ffffff}{16.7} &  \cellCD[22pt]{b7d7a8}{69.8} &  \cellCD[22pt]{69a84f}{90.0} \\
tnt\_intermediate  (8) & 268.6 &  \cellCD[22pt]{ffffff}{5.7e-3} &  \cellCD[22pt]{b7d7a8}{7.8e-5} &  \cellCD[22pt]{69a84f}{1.9e-5} &  \cellCD[22pt]{b7d7a8}{2.6e-4} &  \cellCD[22pt]{ffffff}{89.8} &  \cellCD[22pt]{b7d7a8}{99.9} &  \cellCD[22pt]{69a84f}{100.0} &  \cellCD[22pt]{b7d7a8}{99.8} &  \cellCD[22pt]{ffffff}{61.7} &  \cellCD[22pt]{ffffff}{94.1} &  \cellCD[22pt]{69a84f}{99.0} &  \cellCD[22pt]{b7d7a8}{98.9} &  \cellCD[22pt]{ffffff}{62.8} &  \cellCD[22pt]{b7d7a8}{99.3} &  \cellCD[22pt]{69a84f}{99.9} &  \cellCD[22pt]{b7d7a8}{99.5} &  \cellCD[22pt]{ffffff}{35.4} &  \cellCD[22pt]{ffffff}{83.1} &  \cellCD[22pt]{b7d7a8}{96.9} &  \cellCD[22pt]{69a84f}{97.3} \\
  tnt\_training  (7) & 470.1 &  \cellCD[22pt]{b7d7a8}{3.8e-3} &  \cellCD[22pt]{b7d7a8}{3.0e-3} &  \cellCD[22pt]{ffffff}{1.1e-2} &  \cellCD[22pt]{69a84f}{3.0e-4} &  \cellCD[22pt]{b7d7a8}{88.8} &  \cellCD[22pt]{b7d7a8}{87.8} &  \cellCD[22pt]{b7d7a8}{88.7} &  \cellCD[22pt]{69a84f}{99.9} &  \cellCD[22pt]{ffffff}{63.2} &  \cellCD[22pt]{ffffff}{77.2} &  \cellCD[22pt]{b7d7a8}{87.9} &  \cellCD[22pt]{69a84f}{99.5} &  \cellCD[22pt]{ffffff}{63.6} &  \cellCD[22pt]{ffffff}{82.1} &  \cellCD[22pt]{b7d7a8}{88.6} &  \cellCD[22pt]{69a84f}{99.9} &  \cellCD[22pt]{ffffff}{31.9} &  \cellCD[22pt]{ffffff}{60.5} &  \cellCD[22pt]{b7d7a8}{86.3} &  \cellCD[22pt]{69a84f}{98.7} \\
      nerf\_osr  (8) & 402.8 &  \cellCD[22pt]{b7d7a8}{1.4e-3} &  \cellCD[22pt]{b7d7a8}{1.6e-3} &  \cellCD[22pt]{69a84f}{1.1e-3} &  \cellCD[22pt]{b7d7a8}{1.3e-3} &  \cellCD[22pt]{ffffff}{89.8} &  \cellCD[22pt]{b7d7a8}{91.7} &  \cellCD[22pt]{b7d7a8}{92.0} &  \cellCD[22pt]{69a84f}{92.1} &  \cellCD[22pt]{ffffff}{69.3} &  \cellCD[22pt]{b7d7a8}{70.9} &  \cellCD[22pt]{69a84f}{71.9} &  \cellCD[22pt]{b7d7a8}{71.7} &  \cellCD[22pt]{ffffff}{54.4} &  \cellCD[22pt]{b7d7a8}{71.1} &  \cellCD[22pt]{69a84f}{71.9} &  \cellCD[22pt]{b7d7a8}{71.7} &  \cellCD[22pt]{ffffff}{35.0} &  \cellCD[22pt]{b7d7a8}{43.2} &  \cellCD[22pt]{69a84f}{45.2} &  \cellCD[22pt]{b7d7a8}{44.7} \\
  drone\_deploy  (9) & 524.7 &  \cellCD[22pt]{b7d7a8}{2.4e-3} &  \cellCD[22pt]{b7d7a8}{4.9e-3} &  \cellCD[22pt]{b7d7a8}{4.3e-3} &  \cellCD[22pt]{69a84f}{2.0e-3} &  \cellCD[22pt]{ffffff}{89.6} &  \cellCD[22pt]{b7d7a8}{97.9} &  \cellCD[22pt]{69a84f}{98.2} &  \cellCD[22pt]{ffffff}{91.3} &  \cellCD[22pt]{b7d7a8}{71.8} &  \cellCD[22pt]{b7d7a8}{79.2} &  \cellCD[22pt]{69a84f}{81.1} &  \cellCD[22pt]{ffffff}{65.2} &  \cellCD[22pt]{ffffff}{72.4} &  \cellCD[22pt]{b7d7a8}{89.6} &  \cellCD[22pt]{69a84f}{91.5} &  \cellCD[22pt]{ffffff}{73.5} &  \cellCD[22pt]{b7d7a8}{46.6} &  \cellCD[22pt]{b7d7a8}{50.4} &  \cellCD[22pt]{69a84f}{53.5} &  \cellCD[22pt]{ffffff}{40.2} \\
   urban\_scene  (3) & 3824 &  \cellCD[22pt]{ffffff}{8.1e-5} &  \cellCD[22pt]{b7d7a8}{1.7e-5} &  \cellCD[22pt]{69a84f}{1.4e-5} &  \cellCD[22pt]{69a84f}{1.4e-5} &  \cellCD[22pt]{ffffff}{99.0} &  \cellCD[22pt]{b7d7a8}{99.9} &  \cellCD[22pt]{b7d7a8}{99.9} &  \cellCD[22pt]{69a84f}{100.0} &  \cellCD[22pt]{ffffff}{85.9} &  \cellCD[22pt]{b7d7a8}{95.3} &  \cellCD[22pt]{69a84f}{97.0} &  \cellCD[22pt]{69a84f}{97.0} &  \cellCD[22pt]{ffffff}{94.1} &  \cellCD[22pt]{b7d7a8}{99.5} &  \cellCD[22pt]{69a84f}{99.6} &  \cellCD[22pt]{69a84f}{99.6} &  \cellCD[22pt]{ffffff}{63.2} &  \cellCD[22pt]{ffffff}{86.3} &  \cellCD[22pt]{b7d7a8}{91.2} &  \cellCD[22pt]{69a84f}{91.3} \\
mill19\_building  & 1920 &  \cellCD[22pt]{b7d7a8}{5.1e-5} &  \cellCD[22pt]{b7d7a8}{3.0e-4} &  \cellCD[22pt]{ffffff}{1.3e-2} &  \cellCD[22pt]{69a84f}{1.9e-5} &  \cellCD[22pt]{b7d7a8}{99.6} &  \cellCD[22pt]{69a84f}{99.9} &  \cellCD[22pt]{f4cccc}{0.1} &  \cellCD[22pt]{69a84f}{99.9} &  \cellCD[22pt]{b7d7a8}{93.5} &  \cellCD[22pt]{b7d7a8}{95.5} &  \cellCD[22pt]{f4cccc}{0.0} &  \cellCD[22pt]{69a84f}{95.6} &  \cellCD[22pt]{b7d7a8}{95.9} &  \cellCD[22pt]{69a84f}{99.3} &  \cellCD[22pt]{f4cccc}{0.0} &  \cellCD[22pt]{69a84f}{99.3} &  \cellCD[22pt]{b7d7a8}{81.8} &  \cellCD[22pt]{b7d7a8}{87.0} &  \cellCD[22pt]{f4cccc}{0.0} &  \cellCD[22pt]{69a84f}{87.4} \\
 mill19\_rubble  & 1657 &  \cellCD[22pt]{b7d7a8}{4.3e-5} &  \cellCD[22pt]{b7d7a8}{3.6e-5} &  \cellCD[22pt]{ffffff}{6.4e-5} &  \cellCD[22pt]{69a84f}{3.4e-5} &  \cellCD[22pt]{69a84f}{99.9} &  \cellCD[22pt]{69a84f}{99.9} &  \cellCD[22pt]{b7d7a8}{99.8} &  \cellCD[22pt]{69a84f}{99.9} &  \cellCD[22pt]{69a84f}{95.8} &  \cellCD[22pt]{b7d7a8}{93.6} &  \cellCD[22pt]{b7d7a8}{94.5} &  \cellCD[22pt]{b7d7a8}{94.6} &  \cellCD[22pt]{b7d7a8}{98.4} &  \cellCD[22pt]{b7d7a8}{98.6} &  \cellCD[22pt]{b7d7a8}{98.6} &  \cellCD[22pt]{69a84f}{98.7} &  \cellCD[22pt]{69a84f}{87.8} &  \cellCD[22pt]{ffffff}{81.6} &  \cellCD[22pt]{b7d7a8}{84.7} &  \cellCD[22pt]{b7d7a8}{84.8} \\
eyeful\_apartment  & 3804 &  \cellCD[22pt]{b7d7a8}{3.6e-3} &  \cellCD[22pt]{b7d7a8}{2.8e-3} &  \cellCD[22pt]{ffffff}{9.4e-3} &  \cellCD[22pt]{69a84f}{2.2e-3} &  \cellCD[22pt]{b7d7a8}{57.8} &  \cellCD[22pt]{b7d7a8}{86.8} &  \cellCD[22pt]{ffffff}{75.0} &  \cellCD[22pt]{69a84f}{90.2} &  \cellCD[22pt]{b7d7a8}{34.5} &  \cellCD[22pt]{ffffff}{45.5} &  \cellCD[22pt]{b7d7a8}{50.5} &  \cellCD[22pt]{69a84f}{62.0} &  \cellCD[22pt]{b7d7a8}{21.2} &  \cellCD[22pt]{ffffff}{51.1} &  \cellCD[22pt]{b7d7a8}{61.3} &  \cellCD[22pt]{69a84f}{71.7} &  \cellCD[22pt]{b7d7a8}{8.1} &  \cellCD[22pt]{ffffff}{6.4} &  \cellCD[22pt]{b7d7a8}{18.2} &  \cellCD[22pt]{69a84f}{21.9} \\
eyeful\_kitchen  & 6042 &  \cellCD[22pt]{69a84f}{9.7e-4} &  \cellCD[22pt]{b7d7a8}{3.1e-3} &  \cellCD[22pt]{b7d7a8}{7.4e-3} &  \cellCD[22pt]{c0c0c0}{\vphantom{0}-} &  \cellCD[22pt]{b7d7a8}{72.3} &  \cellCD[22pt]{69a84f}{85.0} &  \cellCD[22pt]{b7d7a8}{59.9} &  \cellCD[22pt]{c0c0c0}{\vphantom{0}-} &  \cellCD[22pt]{69a84f}{46.0} &  \cellCD[22pt]{b7d7a8}{38.1} &  \cellCD[22pt]{b7d7a8}{41.2} &  \cellCD[22pt]{c0c0c0}{\vphantom{0}-} &  \cellCD[22pt]{b7d7a8}{33.7} &  \cellCD[22pt]{b7d7a8}{46.7} &  \cellCD[22pt]{69a84f}{51.7} &  \cellCD[22pt]{c0c0c0}{\vphantom{0}-} &  \cellCD[22pt]{b7d7a8}{13.4} &  \cellCD[22pt]{b7d7a8}{4.6} &  \cellCD[22pt]{69a84f}{14.4} &  \cellCD[22pt]{c0c0c0}{\vphantom{0}-} \\
\bottomrule
\end{tabular}
}
\caption{
\textbf{Pose accuracy comparison} on MipNeRF360~\cite{barron2022mip}, Tanks and Temples~\cite{knapitsch2017tanks}, NeRF-OSR~\cite{rudnev2022nerfosr}, DroneDeploy~\cite{Pilkington2022}, Urbanscene3D~\cite{UrbanScene3D}, Mill-19~\cite{Turki_2022_CVPR}, and Eyeful Tower~\cite{VRNeRF}. We follow the evaluation protocol established by FastMap~\cite{fastmap2025} for reporting pose accuracy metrics on these datasets. 
Note, \cite{fastmap2025} uses a different COLMAP groundtruth to \cref{tab:tnt_pose}. 
For datasets with multiple scenes, we denote the average number of images as \texttt{dataset-name(\#scenes)}. Results are listed separately for each scene in Mill-19 and Eyeful Tower. Metrics are color-coded with {\color{teal}dark green} for best performance and {\color{teal}light green} for competitive performance. {\color{red}Red} denotes complete failures and {\color{gray}gray} indicates timeout (did not finish in one week).
}
\label{tab:fastmap_comparison}
\end{table*}

\subsection{FastMap Benchmark Comparison}
\label{sec:fastmap_benchmark}

Following FastMap~\cite{fastmap2025}, we evaluate our method on a comprehensive set of large-scale  real-world datasets that cover diverse camera trajectory patterns and scene complexities. 
The evaluation includes eight datasets: MipNeRF360~\cite{barron2022mip}, Tanks and Temples~\cite{knapitsch2017tanks}, NeRF-OSR~\cite{rudnev2022nerfosr}, DroneDeploy~\cite{Pilkington2022}, Urbanscene3D~\cite{UrbanScene3D}, Mill-19~\cite{Turki_2022_CVPR}, and Eyeful Tower~\cite{VRNeRF}, with scene sizes ranging from approximately $200$ to $6{,}000$ images per scene. 
We compare against FastMap, GLOMAP~\cite{pan2024glomap}, and COLMAP~\cite{Schonberger_2016_CVPR} using standard pose accuracy metrics including ATE, RTA@$\delta$, and AUC-R\&T@$\delta$ at multiple thresholds.
The results are presented in \cref{tab:fastmap_comparison}, where we report per-dataset averages for multi-scene datasets and individual results for Mill-19 and Eyeful Tower scenes.

\subsection{Ablations}

\Paragraph{Runtime.}
Our work primarily addresses the challenge of applying pre-trained monocular depth estimators (MDE) to multi-view pose estimation.
As a result, computational efficiency has not been a primary focus of this work.
In particular, we used first-order gradient descent for optimization, which can be less efficient than second-order methods.
In \cref{tab:scalability_supp}, we present a runtime comparison with COLMAP.
We report only the Bundle Adjustment time, excluding any preprocessing overhead.
Overall, our method running $50$k iterations is approximately $2\text{--}4\times$ slower than COLMAP.
The use of first-order optimization facilitates scaling up to substantially larger problem set which is nontrivial to achieve with second-order optimization. 
Notably, COLMAP crashes on the Wayspots dataset.
In \cref{tab:optimizer}, we ablate the number of iterations, only running $5$k steps with a sophisticated optimization scheme. 
(Detailed in paragraph Optimization Strategy).
Yet, this requires dataset-specific engineering, hurting the generalization capability of our method.
We leave a more thorough investigation into computational efficiency for future work.

\Paragraph{Number of sampled pixels ($\kappa$).}
Our method requires a certain level of sampling density to achieve the desired level of accuracy, as shown in \cref{tab:optimizer} on ETH3D. 
With insufficient sampling ($\kappa = 30$), the results significantly decrease to $91.7\%$ RRA and $78.6\%$ RTA at $5^\circ$. Our default configuration ($\kappa = 200$) achieves $98.0\%$ RRA and $91.3\%$ RTA. Further increasing the sampling density to $\kappa=500$ yields additional minor improvements, reaching $99.3\%$ RRA and $92.0\%$ RTA, slightly outperforming the numbers reported in the main paper ($97.3\%$ RRA and $90.2\%$ RTA at $5^\circ$ in \cref{tab:eth3d}).

\begin{table*}[t!]
    \centering\footnotesize
    \resizebox{\linewidth}{!}{
    \begin{tabular}{l|cc|cc|cc|cc|cc|cc|cc|cc|cc}
    \toprule
    \multirow{2}{*}{\bf Scene} & \multicolumn{2}{c|}{COLMAP~\cite{Schonberger_2016_CVPR}}& \multicolumn{2}{c|}{ACE-Zero~\cite{brachmann2024acezero}}& \multicolumn{2}{c|}{FlowMap~\cite{smith25flowmap}}& \multicolumn{2}{c|}{VGGSfM~\cite{wang2024vggsfm}}& \multicolumn{2}{c|}{DF-SfM~\cite{He_2024_CVPR}}& \multicolumn{2}{c|}{MASt3R-SfM~\cite{duisterhofmast3rsfm}}&\multicolumn{2}{c|}{Ours / DUSt3R~\cite{Wang_2024_CVPR}}& \multicolumn{2}{c|}{Ours / ZoeDepth~\cite{bhat2023zoedepth}}& \multicolumn{2}{c}{Ours / UniDepth~\cite{piccinelli2024unidepth}}\\
    & RRA & RTA & RRA & RTA & RRA & RTA & RRA & RTA & RRA & RTA & RRA & RTA & RRA & RTA & RRA & RTA & RRA & RTA \\
    \midrule
    courtyard & $56.3$ & $60.0$ & $4.0$ & $1.9$ & $7.5$ & $3.6$ & $50.5$ & $51.2$ & $80.7$ & $74.8$ & $89.8$ & $64.4$ & $94.7$ & $94.7$ & $94.7$ & $94.5$ & $94.7$ & $94.4$ \\
    delivery area & $34.0$ & $28.1$ & $27.4$ & $1.9$ & $29.4$ & $23.8$ & $22.0$ & $19.6$ & $82.5$ & $82.0$ & $83.1$ & $81.8$ & $83.1$ & $83.0$ & $87.8$ & $82.0$ & $83.1$ & $83.1$ \\
    electro & $53.3$ & $48.5$ & $16.9$ & $7.9$ & $2.5$ & $1.2$ & $79.9$ & $58.6$ & $82.8$ & $81.2$ & $100.0$ & $95.5$ & $95.6$ & $78.2$ & $91.9$ & $78.5$ & $93.0$ & $77.2$ \\
    facade & $92.2$ & $90.0$ & $74.5$ & $64.1$ & $15.7$ & $16.8$ & $57.5$ & $48.7$ & $80.9$ & $82.6$ & $74.3$ & $75.3$ & $100.0$ & $99.2$ & $100.0$ & $97.4$ & $80.9$ & $86.0$ \\
    kicker & $87.3$ & $86.2$ & $26.2$ & $16.8$ & $1.5 $ & $1.5$ & $100.0$ & $97.8$ & $93.5$ & $91.0$ & $100.0$ & $100.0$ & $100.0$ & $98.9$ & $100.0$ & $98.5$ & $100.0$ & $98.0$ \\
    meadow & $0.9$ & $0.9$ & $3.8$ & $0.9$ & $3.8$ & $2.9$ & $100.0$ & $96.2$ & $56.2$  & $58.1$ & $58.1$ & $58.1$ & $100.0$ & $58.1$ & $45.7$ & $33.3$ & $100.0$ & $56.7$ \\
    office & $36.9$ & $32.3$ & $0.9$ & $0.0$ & $0.9$ & $1.5$ & $64.9$ & $42.1$ & $71.1$ & $54.5$ & $100.0$ & $98.5$ & $100.0$ & $86.2$ & $100.0$ & $85.7$ & $100.0$ & $86.5$ \\
    pipes & $30.8$ & $28.6$ & $9.9$ & $1.1$ & $6.6$ & $12.1$ & $100.0$ & $97.8$ & $72.5$ & $61.5$ & $100.0$ & $100.0$ & $100.0$ & $96.7$ & $100.0$ & $94.5$ & $100.0$ & $97.8$ \\
    playground & $17.2$ & $18.1$ & $3.8$ & $2.6$ & $2.6$ & $2.8$ & $37.3$ & $40.8$ & $70.5$ & $70.1$ & $100.0$ & $93.6$ & $94.7$ & $93.8$ & $100.0$ & $96.5$ & $100.0$ & $99.2$ \\
    relief & $16.8$ & $16.8$ & $16.8$ & $17.0$ & $6.9$ & $7.7$ & $59.6$ & $57.9$ & $32.9$ & $32.9$ & $34.2$ & $40.2$ & $100.0$ & $98.9$ & $100.0$ & $97.4$ & $100.0$ & $99.6$ \\
    relief 2 & $11.8$ & $11.8$ & $7.3$ & $5.6$ & $8.4$ & $2.8$ & $69.9$ & $70.3$ & $40.9$ & $39.1$ & $57.4$ & $76.1$ & $100.0$ & $98.9$ & $100.0$ & $98.6$ & $100.0$ & $99.8$ \\
    terrace & $100.0$ & $100.0$ & $5.5$ & $2.0$ & $33.2$ & $24.1$ & $38.7$ & $29.6$ &  $100.0$ & $99.6$ & $100.0$ & $100.0$ & $100.0$ & $100.0$ & $100.0$ & $94.5$ & $100.0$ & $98.6$ \\
    terrains & $100.0$ & $99.5$ & $15.8$ & $4.5$ & $12.3$ & $13.8$ & $70.4$ & $54.9$ & $100.0$ & $91.9$ & $58.2$ & $52.5$ & $100.0$ & $95.4$ & $100.0$ & $93.4$ & $100.0$ & $95.2$ \\
    \midrule
    Average & $49.0$ & $47.8$ & $16.4$ & $9.7$ & $10.1$ & $8.8$ & $65.4$ & $58.9$ & $74.2$ & $70.7$ & $81.2$ & $79.7$ & $\mathbf{97.3}$ & $\mathbf{90.2}$ & $93.9$ & $88.1$ & $96.3$ & $\mathbf{90.2}$ \\
    \bottomrule
    \end{tabular}
    }
    \vspace{0ex}
    \caption{\textbf{Structure-from-Motion} ablation with other Monocular Depth Models ZoeDepth and UniDepth on ETH3D dataset~\cite{Schops_2017_CVPR, Schops_2019_CVPR}.}
    \label{tab:eth3d_plus}
\end{table*}

\begin{table*}[t]
    \centering\small
    \resizebox{0.8\linewidth}{!}{%
    \begin{tabular}{l|l|l|ccc|ccc}
    \toprule
    \multirow{3}{*}{\bf Method} & \multirow{3}{*}{Depth} & \multirow{3}{*}{Corres.} & \multicolumn{3}{c|}{Calibrated} & \multicolumn{3}{c}{Uncalibrated} \\
    &&&$\text{Acc}@3^\circ$&$\text{Acc}@5^\circ$&$\text{Acc}@10^\circ$&$\text{Acc}@3^\circ$&$\text{Acc}@5^\circ$&$\text{Acc}@10^\circ$\\ 
    \midrule
    COLMAP~\cite{Schonberger_2016_CVPR}&- & SuperPoint~\cite{detone2018superpoint} & ${0.398}$ & $0.589$ & $0.783$ & ${0.342}$ & ${0.505}$ & $0.670$ \\
    \midrule
    \multirow{5}{*}{MBA (Ours)}
    & ZoeDepth~\cite{bhat2023zoedepth} & RoMa~\cite{edstedt2024roma} & $0.396$ & ${0.614}$ & ${0.823}$ & ${0.372}$ & ${0.586}$ & ${0.811}$ \\
    & DUSt3R~\cite{Wang_2024_CVPR}  & RoMa~\cite{edstedt2024roma} & $0.426$ & $0.631$ & $0.830$ & $0.403$ & $\mathbf{0.615}$ & $0.820$ \\
    & UniDepth~\cite{piccinelli2024unidepth} & RoMa~\cite{edstedt2024roma} & $0.432$ & $0.636$ & $0.833$ & $\mathbf{0.407}$ & $0.612$ & $\mathbf{0.823}$ \\
    & DUSt3R~\cite{Wang_2024_CVPR}  & MASt3R~\cite{mast3r_arxiv24} & $0.432$ & $0.639$ & $0.837$ & $0.384$ & $0.596$ & $0.811$ \\
    & UniDepth~\cite{piccinelli2024unidepth} & MASt3R~\cite{mast3r_arxiv24} & $\mathbf{0.439}$ & $\mathbf{0.645}$ & $\mathbf{0.841}$ & $0.393$ & $0.598$ & $0.817$ \\
    \bottomrule
    \end{tabular}
    }
    \caption{
    \textbf{Extended Structure-from-Motion Results} on the ScanNet dataset~\cite{ScanNet_Dai_2017_CVPR}, with calibrated and uncalibrated cases.
    }
    \label{tab:scannet_full}
\end{table*}

\Paragraph{Loss Function Comparison.}
We ablate different losses in \cref{tab:optimizer} for their effectiveness for monocular depth-based pose estimation. 
Our marginalized Bundle-Adjustment (MBA) approach significantly outperforms traditional loss functions, achieving $97.3\%$ RRA and $90.2\%$ RTA. 
Among conventional losses, the soft L1 ($87.5\%$ RRA, $73.9\%$ RTA) and Cauchy ($86.2\%$ RRA, $75.2\%$ RTA) show moderate performance, while the l2 loss performs worst ($79.4\%$ RRA, $66.0\%$ RTA). 
These findings highlight the importance of robust loss functions that can effectively handle noise in monocular depth estimates.

\subsection{Extended ScanNet and 7-Scenes Results}

We also report results on the ScanNet dataset using calibrated cameras (\textit{i.e.}, known intrinsics) in \cref{tab:scannet_full}. 
We further present the detailed per-sequence performance on the 7-Scenes dataset in \cref{tab:seven_scenes_full}.

\subsection{Evaluation Protocols}
\label{sec:protocol}
\Paragraph{ETH3D Dataset in \cref{tab:eth3d}.}
The ETH3D consists of $13$ multi-view scenes containing up to $76$ high-resolution photographs per scene. 
We evaluate on ETH3D dataset following two evaluation protocols.
\cref{tab:eth3d} follows MASt3R-SfM~\cite{duisterhofmast3rsfm} to report \textit{Relative Rotation Accuracy} (RRA@$\tau$) and \textit{Relative Translation Accuracy} (RTA@$\tau$), which measure the percentage of image pairs whose estimated relative pose errors fall below a threshold $\tau=5^\circ$. For each image pair $(i, j)$ with valid ground-truth poses, the rotation error is computed as the angular difference between the estimated and ground-truth relative rotations, \textit{i.e.}, the angle of $R_{ij}^{\text{gt}^{-1}} R_{ij}^{\text{est}}$, while the translation error is the angle between the normalized translation directions $t_{ij}^{\text{gt}}$ and $t_{ij}^{\text{est}}$. These errors are aggregated over all possible image pairs within each scene, yielding one RRA and one RTA score per scene. The final reported scores are obtained by averaging the per-scene RRA and RTA across all $13$ scenes.

\Paragraph{IMC2021 in \cref{tab:imc}}
We follow DF-SfM~\cite{He_2024_CVPR} in evaluating the Area Under the Curve (AUC) of relative pose accuracy at multiple thresholds. For each image pair with ground-truth camera poses, we compute the rotation error as the angular difference (in degrees) between the estimated and ground-truth relative rotations, and the translation error as the angle between the corresponding normalized translation directions. The pose error is defined as the maximum of the rotation and translation errors. The AUC at threshold $\tau$ is defined as the area under the cumulative distribution function (CDF) of pose errors up to $\tau$, normalized by $\tau$, where $\text{CDF}(e)$ denotes the proportion of image pairs with pose error less than $e$ degrees. 
On the IMC dataset, the AUC metric is computed globally across all valid image pairs, without per-scene aggregation. Following standard practice, we report AUC at thresholds: $3^\circ$, $5^\circ$, and $10^\circ$.

\Paragraph{ScanNet Dataset.}
Similarly, for the ScanNet dataset, we define the pose error as the angular error given by the maximum of the rotation and translation angular errors. 
For each image pair with ground-truth poses, the rotation error is computed as the angular difference between the estimated and ground-truth relative rotations, and the translation error is defined as the angle between the corresponding normalized translation vectors. 
We report the \textit{Accuracy} (ACC@$\tau$), defined as the percentage of image pairs with pose error less than a given threshold $\tau$. 
The metric is computed over all visible frame pairs within each scene and averaged to produce a per-scene ACC score. 
The final result is average across all scenes. 
For a fair comparison with COLMAP, we only report scores over the subset of frame pairs for which COLMAP successfully returns a pose estimate.


\Paragraph{FastMap Benchmark.}
Following FastMap~\cite{fastmap2025}, we report standard pose accuracy metrics including Absolute Translation Error (ATE), Relative Translation Accuracy (RTA@$\delta$), and Area Under the Curve for Rotation and Translation (AUC-R\&T@$\delta$) at thresholds of $1^\circ$ and $3^\circ$. For each image pair with valid ground-truth poses, the rotation error is computed as the angular difference between estimated and ground-truth relative rotations, and the translation error as the angle between normalized translation directions. ATE measures the absolute translation error magnitude. RTA@$\delta$ reports the percentage of image pairs with both rotation and translation errors below threshold $\delta$. AUC-R\&T@$\delta$ computes the area under the accuracy curve up to threshold $\delta$. For multi-scene datasets, we report per-dataset averages, while Mill-19 and Eyeful Tower results are listed per scene.

\begin{table*}[t!]
    \centering
    \small
    \resizebox{0.9\linewidth}{!}{%
    \begin{tabular}{ll|ccccccc|c}
    \toprule
    {\bf Category} & {\bf Method}
     & Chess & Fire & Heads & Office & Pumpkin & Kitchen & Stairs & Average\\
    \midrule
    \multirow{2}{*}{FM} & AS~\cite{sattler2016efficient}\tiny{PAMI'16} & $4/1.96$ & $3/1.53$ & $2/1.45$ & $9/3.61$ & $8/3.10$ &  $7/3.37$ & $3/2.22$ & $5.1/2.46$\\
    &HLoc~\cite{sarlin2019coarse}\tiny{CVPR'19} & $2/0.79$ & $2/0.87$ & $2/0.92$ & $3/0.91$ & $5/1.12$ & $4/1.25$ & $6/1.62$ & $3.4/1.07$ \\
    \midrule
    \multirow{3}{*}{E2E}
    & SC-wLS~\cite{wu2022sc}\tiny{ECCV'22} &  $\rx{3}/0.76$ & $\rx{5}/1.09$ & $\rx{3}/1.92$ & $\rx{6}/0.86$ & $\rx{8}/1.27$ & $\rx{9}/1.43$ & $\rx{12}/2.80$ & $6.6/1.45$ \\
    & NeuMaps~\cite{Tang_2023_CVPR}\tiny{CVPR'23} & $\rx{2}/0.81$ & $\rx{3}/1.11$ & $\rx{2}/1.17$ & $\rx{3}/0.98$ & $\rx{4}/1.11$ & $\rx{4}/1.33$ & $\rx{4}/1.12$ & $3.1/1.09$ \\
    & PixLoc~\cite{Sarlin_2021_CVPR}\tiny{CVPR'21} & $\rx{2}/0.80$ & $\rx{2}/0.73$ & $\rx{1}/0.82$ & $\rx{3}/0.82$ & $\rx{4}/1.21$ & $\rx{3}/1.20$ & $\rx{5}/1.30$ & $2.9/0.98$\\
    \midrule
    \multirow{4}{*}{SCR}
    & ACE~\cite{Brachmann_2023_CVPR}\tiny{CVPR'23} & $1.9/0.7$ & $1.9/0.9$ & $0.9/0.6$ & $2.7/0.8$ & $4.2/1.1$ & $4.2/1.3$ & $3.9/1.1$ & $2.8/0.93$ \\
    & DSAC*~\cite{DSAC_PAMI2022}\tiny{PAMI'22} & $1.9/1.11$ & $1.9/1.24$ & $1.1/1.82$ & $2.6/1.18$ & $4.2/1.41$ & $3.0/1.70$ & $4.2/1.42$ & $2.7/1.41$\\
    & HSCNet~\cite{Li_2020_CVPR}\tiny{CVPR'20} & $\rx{2}/0.7$ & $\rx{2}/0.9$ & $\rx{1}/0.9$ & $\rx{3}/0.8$ & $\rx{4}/1.0$ & $\rx{4}/1.2$ & $\rx{3}/0.8$ & $2.7/0.90$\\
    & HSCNet++~\cite{Wang2024}\tiny{IJCV'24} & $\rx{2}/0.63$ & $\rx{2}/0.79$ & $\rx{1}/0.8$ & $\rx{2}/0.65$ & $\rx{3}/0.85$ & $\rx{3}/1.09$ & $\rx{3}/0.83$ & $2.29/0.81$\\
    \midrule
    \multirow{3}{*}{APR}
    & Direct-PN~\cite{DirectPN_2021_3DV}\tiny{3DV'21} & $10/3.52$ & $27/8.66$ & $17/13.1$ & $16/5.96$ & $19/3.85$ & $22/5.13$ & $32/10.6$ & $20/7.26$ \\
    & DFNet~\cite{chen2022dfnet}\tiny{ECCV'22} & $3/1.15$ & $9/3.71$ & $8/6.08$ & $7/2.14$ & $10/2.76$ & $9/2.87$ & $11/5.58$ & $8/3.47$\\
    & MAREPO~\cite{marepo_Chen_2024_CVPR}\tiny{CVPR'24} & $2.1/1.24$ & $2.3/1.39$ & $1.8/2.03$ & $2.8/1.26$ & $3.5/1.48$ & $4.2/1.71$ & $5.6/1.67$ & $3.2/1.54$ \\
    \midrule
    MDE & MBA (Ours) & $2.2/0.77$ & $1.9/0.80$ & $1.1/0.80$ & $3.0/0.91$ & $4.3/1.04$ & $3.7/1.32$ & $2.7/0.78$ & $2.7/0.92$ \\
    \bottomrule
    \end{tabular}
    }
    \caption{\textbf{Extended Camera Relocalization Results} on the 7-Scenes dataset~\cite{shotton2013scene}, with per-scene performance.}
    \label{tab:seven_scenes_full}
\end{table*}

\Paragraph{7-Scenes Dataset.}
For the 7-Scenes dataset~\cite{shotton2013scene}, we follow DUSt3R~\cite{Wang_2024_CVPR} on the standard evaluation protocol by computing the median rotation and translation errors across all test frames. 
The translation error is measured as the Euclidean distance (in centimeters) between the predicted and ground-truth camera positions, while the rotation error is computed as the angular difference (in degrees) between the predicted and ground-truth orientations. 
These median errors provide a robust summary of pose estimation accuracy in the presence of outliers and are reported for each scene individually. The final scores are obtained by averaging the per-scene median errors across all seven scenes.

\Paragraph{Wayspots Dataset.}
For the Wayspots dataset~\cite{Brachmann_2023_CVPR}, we follow the official evaluation protocol by measuring the accuracy of absolute camera pose predictions at multiple error thresholds. Specifically, the predicted pose is considered correct if its translation error is below $10$ cm and its rotation error is below $5^\circ$, computed with respect to the ground-truth camera pose. For each test sequence, we report the percentage of query images that meet this criterion. The final performance is obtained by averaging the per-sequence accuracies across all test scenes.

\subsection{Extended Implementation Details}
\label{sec:supp_details}
In both coarse and fine stages, we optimize with Adam~\cite{kingma2014adam} for $50,000$ iterations at a learning rate of 1e-3. 
Within each pair of frames, we sample $\kappa = 200$ pixels. 
For coarse BA objective \cref{eqn:coarse_ba}, we set maximum logged residual value $\bar{\tau}_{\text{max}} = 10$.
For fine BA objective \cref{eqn:binary_scoring_funciton}, we set $\tau_{\text{max}} = 20$.
We parameterize camera poses following SPARF~\cite{truong2023sparf}, where rotations are represented using a $6$-DoF continuous representation and translations are encoded as $3$-DoF vectors. 
We include image pairs where at least $\nu \geq 15\%$ of the pixels are co-visible. 
During preprocessing, we sample correspondences only from dense regions where the confidence score exceeds a threshold of $\chi > 0.2$. 
To compute the probability density function (PDF) and cumulative distribution function (CDF) using a histogram-based kernel density estimation (KDE) algorithm, we use a $1 \times 100$ histogram vector. 
For multi-GPU parallelization, we adopt different strategies in the coarse and fine stages. 
In the coarse stage, we distribute multiple complete sub-graphs (as shown in \cref{fig:pipeline}) across different GPUs. 
In the fine stage, we randomly assign frame pairs to different GPUs for processing. 
We observe that the intrinsic parameters typically converge more slowly than the others; therefore, we increase their learning rate by a factor of $50$, \textit{i.e.}, the intrinsic parameters use a learning rate of 5e-2.

\Paragraph{Two-View Pose Initialization.}
We initialize the two-view pose sequentially. First, we estimate the essential matrix using the five-point algorithm~\cite{li2006five} within a RANSAC loop on normalized image coordinates, and decompose it to recover the rotation \(\mathbf{R}\) and the unit translation direction \(\hat{\mathbf{t}}\) with \(\|\hat{\mathbf{t}}\| = 1\). Next, we resolve the absolute translation scale using the monocular depth map \(\mathbf{D}_i\) of the source frame \(I_i\). For each pixel \(p \in I_i\) with depth \(d(p) = \mathbf{D}_i[p]\), we use the projection operator \(\pi_{i \to j}\) to project it to the target frame \(I_j\) for a candidate scale \(s > 0\). The projection \(\pi_{i \to j}(d(p), s)\) depends on the camera intrinsics \(\mathbf{K}_i\), \(\mathbf{K}_j\) and the relative pose \(\mathbf{R}, s\hat{\mathbf{t}}\). We choose \(s(p)\) that minimizes the distance between the projected point \(\pi_{i \to j}(d(p), s)\) and the corresponding pixel \(v \in I_j\) along the epipolar line defined by \((I_i, I_j)\). Repeating this process for all valid pixels yields a set of per-pixel scale estimates \(\{ s(p) \}\). The median of these estimates is taken as the final scale \(s^\star\), forming the initialized two-view pose \(\left( \mathbf{R}, \; s^\star \hat{\mathbf{t}} \right)\).

\Paragraph{Camera Intrinsic Initialization.}
In uncalibrated settings, we initialize camera intrinsics using the DUSt3R~\cite{Wang_2024_CVPR} pointmap. First, we convert the pointmap into an incidence field following WildCamera~\cite{zhu2023tame}: for each pixel $\mathbf{p}$ with corresponding 3D point $(x,y,z)$, we compute the incidence (ray-direction) vector $(x/z,y/z,1)$. This incidence map encodes the incoming camera-ray direction for each pixel, which, under the pinhole model, depends solely on camera intrinsics and pixel coordinates. We then apply WildCamera's RANSAC-based calibration procedure to recover the $1$ DoF intrinsic. We repeat the intrinsic initialization process for each frame. When shared intrinsics are assumed across the collection, we set the initial focal length to the median value across all frames.

\end{document}